\documentclass{article}

    \usepackage[nonatbib,preprint]{neurips_2021}

\usepackage[utf8]{inputenc} 
\usepackage[T1]{fontenc}    
\usepackage{hyperref}       
\usepackage{url}            
\usepackage{booktabs}       
\usepackage{amsfonts}       
\usepackage{nicefrac}       
\usepackage{microtype}      
\usepackage{xcolor}         

\usepackage{amsfonts}
\usepackage{amsmath}
\usepackage{diagbox}
\usepackage{graphicx}
\usepackage{multicol}
\usepackage{multirow}
\usepackage{url}

\makeatletter
\renewcommand{\vec}[1]{\mathbf{#1}}
\newcommand{\E}{\mathbb{E}}
\makeatother
\newcommand{\orcid}[1]{\href{https://orcid.org/#1}{\includegraphics[height=10pt]{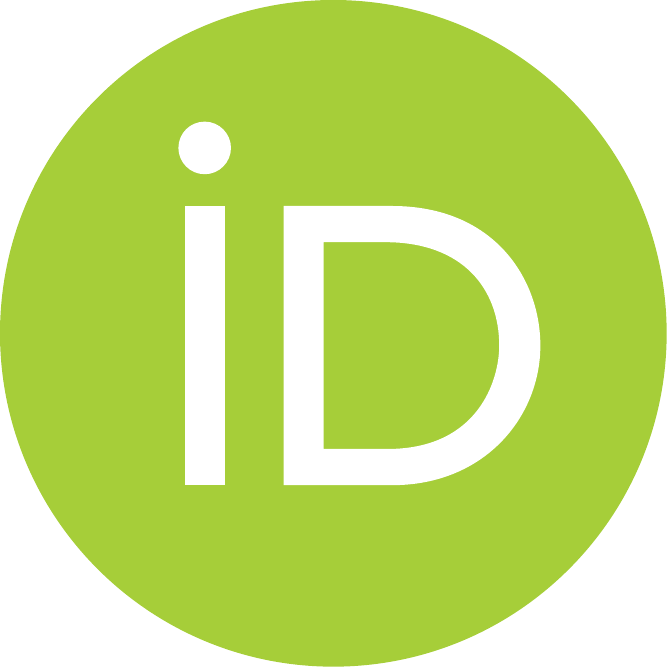}}}
\usepackage[%
style=apa,%
backend=biber,%
citestyle=authoryear,%
natbib=true,%
uniquename=false,%
useprefix=true%
]{biblatex}
\title{Does the Adam Optimizer Exacerbate\\Catastrophic Forgetting?}

\author{%
  Dylan R.~Ashley \orcid{0000-0001-6148-8802} \thanks{Corresponding author} \\
  The Swiss AI Lab IDSIA/USI/SUPSI \\
  Lugano, Ticino, Switzerland \\
  \texttt{dylan.ashley@idsia.ch} \\
  \And
  Sina Ghiassian,\enspace Richard S.~Sutton \\
  Department of Computing Science \\
  University of Alberta \\
  Edmonton, Alberta, Canada \\
  \texttt{\{ghiassia, rsutton\}@ualberta.ca} \\
}

\begin{document}

\maketitle

\begin{abstract}

Catastrophic forgetting remains a severe hindrance to the broad application of artificial neural networks (ANNs), however, it continues to be a poorly understood phenomenon. Despite the extensive amount of work on catastrophic forgetting, we argue that it is still unclear how exactly the phenomenon should be quantified, and, moreover, to what degree all of the choices we make when designing learning systems affect the amount of catastrophic forgetting. We use various testbeds from the reinforcement learning and supervised learning literature to (1)~provide evidence that the choice of which modern gradient-based optimization algorithm is used to train an ANN has a significant impact on the amount of catastrophic forgetting and show that---surprisingly---in many instances classical algorithms such as vanilla SGD experience less catastrophic forgetting than the more modern algorithms such as Adam. We empirically compare four different existing metrics for quantifying catastrophic forgetting and (2)~show that the degree to which the learning systems experience catastrophic forgetting is sufficiently sensitive to the metric used that a change from one principled metric to another is enough to change the conclusions of a study dramatically. Our results suggest that a much more rigorous experimental methodology is required when looking at catastrophic forgetting. Based on our results, we recommend inter-task forgetting in supervised learning must be measured with both retention and relearning metrics concurrently, and intra-task forgetting in reinforcement learning must---at the very least---be measured with pairwise interference.

\end{abstract}

\section{Introduction}
\label{sec:introduction}

In online learning, catastrophic forgetting refers to the tendency for artificial neural networks (ANNs) to forget previously learned information when in the presence of new information~\citep[p.~173]{french1991using}. Catastrophic forgetting presents a severe issue for the broad applicability of ANNs as many important learning problems, such as reinforcement learning, are online learning problems. Efficient online learning is also core to the continual---sometimes called lifelong~\citep[p.~55]{chen2018lifelong}---learning problem.

The existence of catastrophic forgetting is of particular relevance now as ANNs have been responsible for a number of major artificial intelligence (AI) successes in recent years (e.g., \citet{taigman2014deepface}, \citet{mnih2015human}, \citet{silver2016mastering}, \citet{gatys2016image}, \citet{vaswani2017attention}, \citet{radford2019language}, \citet{senior2020improved}). Thus there is reason to believe that methods able to successfully mitigate catastrophic forgetting could lead to new breakthroughs in online learning problems.

The significance of the catastrophic forgetting problem means that it has attracted much attention from the AI community. It was first formally reported on in \citet{mccloskey1989catastrophic} and, since then, numerous methods have been proposed to mitigate it (e.g., \citet{kirkpatrick2017overcoming}, \citet{lee2017overcoming}, \citet{zenke2017continual}, \citet{masse2018alleviating}, \citet{sodhani2020toward}). Despite this, it continues to be an unsolved issue~\citep{kemker2018measuring}. This may be partly because the phenomenon itself---and what contributes to it---is poorly understood, with recent work still uncovering fundamental connections (e.g., \citet{mirzadeh2020understanding}).

This paper is offered as a step forward in our understanding of the phenomenon of catastrophic forgetting. In this work, we seek to improve our understanding of it by revisiting the fundamental questions of (1)~how we should quantify catastrophic forgetting, and (2)~to what degree do all of the choices we make when designing learning systems affect the amount of catastrophic forgetting. To answer the first question, we compare several different existing measures for catastrophic forgetting: retention, relearning, activation overlap, and pairwise interference. We discuss each of these metrics in detail in Section~\ref{sec:measuring_catastrophic_forgetting}. We show that, despite each of these metrics providing a principled measure of catastrophic forgetting, the relative ranking of algorithms varies wildly between them. This result suggests that catastrophic forgetting is not a phenomenon that a single one of these metrics can effectively describe. As most existing research into methods to mitigate catastrophic forgetting rarely looks at more than one of these metrics, our results imply that a more rigorous experimental methodology is required in the research community.

Based on our results, we recommend that work looking at inter-task forgetting in supervised learning must, at the very least, consider both retention and relearning metrics concurrently. For intra-task forgetting in reinforcement learning, our results suggest that pairwise interference may be a suitable metric, but that activation overlap should, in general, be avoided as a singular measure of catastrophic forgetting.

To address the question of to what degree all the choices we make when designing learning systems affect the amount of catastrophic forgetting, we look at how the choice of which modern gradient-based optimizer is used to train an ANN impacts the amount of catastrophic forgetting that occurs during training. We empirically compare vanilla SGD, SGD with Momentum~\citep{qian1999momentum,rumelhart1986learning}, RMSProp~\citep{geoffrey-rmsprop}, and Adam~\citep{kingma2014adam}, under the different metrics and testbeds. Our results suggest that selecting one of these optimizers over another does indeed result in a significant change in the catastrophic forgetting experienced by the learning system. Furthermore, our results ground previous observations about why vanilla SGD is often favoured in continual learning settings~\citep[p.~6]{mirzadeh2020understanding}: namely that it frequently experiences less catastrophic forgetting than the more sophisticated gradient-based optimizers---with a particularly pronounced reduction when compared with Adam. To the best of our knowledge, this is the first work explicitly providing strong evidence of this.

Importantly, in this work, we are trying to better understand the phenomenon of catastrophic forgetting itself, and not explicitly seeking to understand the relationship between catastrophic forgetting and performance. While that relation is important, it is not the focus of this work. Thus, we defer all discussion of that relation to Appendix~\ref{app:actual_performance_on_the_testbeds} of our supplementary material. The source code for our experiments is available at \url{https://github.com/dylanashley/catastrophic-forgetting/tree/arxiv}.

\section{Related Work}
\label{sec:related_work}

This section connects several closely related works to our own and examines how our work compliments them. The first of these related works, \citet{kemker2018measuring}, directly observed how different datasets and different metrics changed the effectiveness of contemporary algorithms designed to mitigate catastrophic forgetting. Our work extends their conclusions to non-retention-based metrics and to more closely related algorithms. \citet{hetherington1989catastrophic} demonstrated that the severity of the catastrophic forgetting shown in the experiments of \citet{mccloskey1989catastrophic} was reduced if catastrophic forgetting was measured with relearning-based rather than retention-based metrics. Our work extends their ideas to more families of metrics and a more modern experimental setting. \citet{goodfellow2013empirical} looked at how different activation functions affected catastrophic forgetting and whether or not dropout could be used to reduce its severity. Our work extends their work to the choice of optimizer and the metric used to quantify catastrophic forgetting.

While we provide the first formal comparison of modern gradient-based optimizers with respect to the amount of catastrophic forgetting they experience, others have previously hypothesized that there could be a potential relation. \citet{ratcliff1990connectionist} contemplated the effect of momentum on their classic results around catastrophic forgetting and then briefly experimented to confirm their conclusions applied under both SGD and SGD with Momentum. While they only viewed small differences, our work demonstrates that a more thorough experiment reveals a much more pronounced effect of the optimizer on the degree of catastrophic forgetting. Furthermore, our work includes the even more modern gradient-based optimizers in our comparison (i.e., RMSProp and Adam), which---as noted by \citet[p.~6]{mirzadeh2020understanding}---are oddly absent from many contemporary learning systems designed to mitigate catastrophic forgetting.

\section{Problem Formulation}
\label{sec:problem_formulation}

In this section, we define the two problem formulations we will be considering in this work. These problem formulations are online supervised learning and online state value estimation in undiscounted, episodic reinforcement learning.

The supervised learning task is to learn a mapping \(f: \mathbb{R}^n \rightarrow \mathbb{R}\) from a set of examples \((\vec{x}_0, y_0)\), \((\vec{x}_1, y_1)\), ..., \((\vec{x}_n, y_n)\). The supervised learning framework is a general one as each $\vec{x_i}$ could be anything from an image to the full text of a book, and each $y_i$ could be anything from the name of an animal to the average amount of time needed to read something. In the incremental online variant of supervised learning, each example \((\vec{x}_t, y_t)\) only becomes available to the learning system at time $t$ and the learning system is expected to learn from only this example at time $t$.

Reinforcement learning considers an agent interacting with an environment. Often this is formulated as a Markov Decision Process, where, at each time step $t$, the agent observes the current state of the environment \(S_{t} \in \mathcal{S}\), takes an action \(A_{t} \in \mathcal{A}\), and, for having taken action \(A_{t}\) when the environment is in state \(S_{t}\), subsequently receives a reward \(R_{t + 1} \in \mathbb{R}\). In episodic reinforcement learning, this continues until the agent reaches a terminal state \(S_{T} \in \mathcal{T} \subset \mathcal{S}\). In undiscounted policy evaluation in reinforcement learning, the goal is to learn, for each state, the expected sum of rewards received before the episode terminates when following a given policy~\citep[p.~74]{sutton2018reinforcement}. Formally we write this as:
\[
    \forall s \in \mathcal{S}, v_{\pi}(s) := \E_{\pi}\left[\sum_{t = 0}^{T} R_{t} | S_{0} = s\right]
\]
where $\pi$ is the policy mapping states to actions, and $T$ is the number of steps left in the episode. We refer to $v_{\pi}(s)$ as the value of state $s$ under policy $\pi$. In the incremental online variant of value estimation in undiscounted episodic reinforcement learning, each transition \((S_{t - 1}, R_{t}, S_{t})\) only becomes available to the learning system at time $t$ and the learning system is expected to learn from only this transition at time $t$.

\section{Measuring Catastrophic Forgetting}
\label{sec:measuring_catastrophic_forgetting}

In this section, we examine the various ways which people have proposed to measure catastrophic forgetting. The most prominent of these is \textit{retention}. Retention-based metrics directly measure the drop in performance on a set of previously-learned tasks after learning a new task. Retention has its roots in psychology (e.g., \citet{barnes1959fate}), and \citet{mccloskey1989catastrophic} used this as a measure of catastrophic forgetting. The simplest way of measuring the retention of a learning system is to train it on one task until it has mastered that task, then train it on a second task until it has mastered that second task, and then, finally, report the new performance on the first task. \citet{mccloskey1989catastrophic} used it in a two-task setting, but more complicated formulations exist for situations where there are more than two tasks (e.g., see \citet{kemker2018measuring}).

An alternative to retention that likewise appears in psychological literature and the machine learning literature is \textit{relearning}. Relearning was the first formal metric used to quantify forgetting in the psychology community~\citep{ebbinghaus1885memory}, and was first used to measure catastrophic forgetting in \citet{hetherington1989catastrophic}. The simplest way of measuring relearning is to train a learning system on a first task to mastery, then train it on a second task to mastery, then train it on the first task to mastery again, and then, finally, report how much quicker the learning system mastered the first task the second time around versus the first time. While in some problems relearning is of lesser import than retention, in others it is much more significant. A simple example of such a problem is one where forgetting is made inevitable due to resource limitations, and the rapid reacquisition of knowledge is paramount.

A third measure for catastrophic forgetting, \textit{activation overlap}, was introduced in \citet{french1991using}. In that work, \citeauthor{french1991using} argued that catastrophic forgetting was a direct consequence of the overlap of the distributed representations of ANNs. He then postulated that catastrophic forgetting could be measured by quantifying the degree of this overlap exhibited by the ANN. The original formulation of the activation overlap of an ANN given a pair of samples looks at the activation of the hidden units in the ANN and measures the element-wise minimum of this between the samples. To bring this idea in line with contemporary thinking (e.g., \citet{kornblith2019similarity}) and modern network design, we propose instead using the dot product of these activations between the samples. Mathematically, we can thus write the activation overlap of a network with hidden units $h_0$, $h_1$, ..., $h_n$ with respect to two samples $\vec{a}$ and $\vec{b}$ as
\[
    s(\vec{a}, \vec{b}) := \frac{1}{n} \sum\limits_{i=0}^{n} g_{h_i} (\vec{a}) \cdot g_{h_i} (\vec{b})
\]
where $g_{h_i} (\vec{x})$ is the activation of the hidden unit $h_i$ with a network input $\vec{x}$.

A more contemporary measure of catastrophic forgetting than activation overlap is \textit{pairwise interference}~\citep{riemer2019learning,liu2019sparse,ghiassian2020improving}. Pairwise interference seeks to explicitly measure how much a network learning from one sample interferes with learning on another sample. In this way, it corresponds to the tendency for a network---under its current weights---to demonstrate both positive transfer and catastrophic forgetting due to interference. Mathematically, the pairwise interference of a network for two samples $\vec{a}$ and $\vec{b}$ at some instant $t$ can be written as
\[
    PI(\theta_t; \vec{a}, \vec{b}) := J(\theta_{t + 1}; \vec{a}) - J(\theta_{t}; \vec{a})
\]
where $J(\theta_t; \vec{a})$ is the performance of the learning system with parameters $\theta_t$ on the objective function $J$ for $\vec{a}$ and $J(\theta_{t + 1}; \vec{a})$ is the performance on $J$ for $\vec{a}$ after performing an update at time $t$ using $\vec{b}$ as input. Assuming $J$ is a measure of error that the learning system is trying to minimize, lower values of pairwise interference suggest that less catastrophic forgetting is occurring.

When comparing the above metrics, note that, unlike activation overlap and pairwise interference, retention and relearning require some explicit notion of ``mastery'' for a given task. Furthermore, note that activation overlap and pairwise interference can be reported at each step during the learning of a single task and thus can measure intra-task catastrophic forgetting, whereas retention and relearning can only measure inter-task catastrophic forgetting. Finally, note that activation overlap and pairwise interference are defined for pairs of samples, whereas retention and relearning are defined over an entire setting. Setting-wide variants of activation overlap and pairwise interference are estimated by just obtaining an average value for them between all pairs in some preselected set of examples.

\section{Experimental Setup}
\label{sec:experimental_setup}

In this section, we design the experiments which will help answer our earlier questions: (1)~how we should quantify catastrophic forgetting, and (2)~to what degree do all of the choices we make when designing learning systems affect the amount of catastrophic forgetting. To address these questions, we apply the four metrics from the previous section to four different testbeds. For brevity, we defer less relevant details of our experimental setup to Appendix~\ref{app:additional_experimental_setup} of our supplementary material.

The first two testbeds we use build on the MNIST~\citep{lecun1998gradient} and Fashion MNIST~\citep{xiao2017fashion} dataset, respectively, to create two four-class image classification supervised learning tasks. With both tasks, we separate the overall task into two distinct subtasks where the first subtask only includes examples from the first and second class, and the second subtask only includes examples from the third and fourth class. We have the learning system learn these subtasks in four phases, wherein only the first and third phases contain the first subtask, and only the second and fourth phases contain the second subtask. Each phase transitions to the next only when the learning system has achieved mastery in the phase. Here, that means the learning system must maintain a running accuracy in that phase of $90\%$ for five consecutive steps. All learning here---and in the other two testbeds---is fully online and incremental.

The third and fourth testbeds draw examples from an agent operating under a fixed policy in a standard undiscounted episodic reinforcement learning domain. For the third testbed, we use the Mountain Car domain~\citep{moore1990efficient,sutton1998reinforcement} and, for the fourth testbed, we use the Acrobot domain~\citep{sutton1995generalization,dejong1994swinging,spong1989robot}. The learning system's goal in both testbeds is to learn, for each timestep, what the value of the current state is. Thus these are both reinforcement learning value estimation testbeds~\citep[p.~74]{sutton2018reinforcement}.

There are several significant differences between the four testbeds that are worth noting. Firstly, the MNIST and Fashion MNIST testbeds' data-streams consist of multiple phases, each containing only i.i.d. examples. However, the Mountain Car and Acrobot testbeds have only one phase each, and that phase contains strongly temporally-correlated examples. One consequence of this difference is that only intra-task catastrophic forgetting metrics can be used in the Mountain Car and Acrobot testbed, and so here, the retention and relearning metrics of Section~\ref{sec:measuring_catastrophic_forgetting} can only be measured in the MNIST and Fashion MNIST testbeds. While it is theoretically possible to derive semantically similar metrics for the Mountain Car and Acrobot testbeds, this is non-trivial as---in addition to them consisting of only a single phase---it is somewhat unclear what mastery is in these contexts. Another difference between the MNIST testbed and the other two testbeds is that in the MNIST testbed---since the network is solving a four-class image classification problem in four phases with not all digits appearing in each phase---some weights connected to the output units of the network will be protected from being modified in some phases. This property of these kinds of experimental testbeds has been noted previously in \citet[Section 6.3.2.]{farquhar2018towards}. In the Mountain Car and Acrobot testbeds, no such weight protection exists.

For each of the four testbeds, we use feedforward shallow ANNs trained through backpropagation~\citep{rumelhart1986learning}. We experiment with four different optimizers for training the aforementioned ANNs for each of the four testbeds. These optimizers are (1)~SGD, (2)~SGD with Momentum~\citep{qian1999momentum,rumelhart1986learning}, (3)~RMSProp~\citep{geoffrey-rmsprop}, and (4)~Adam~\citep{kingma2014adam}. For Adam, in accordance with recommendations of Adam's creators~\citep{kingma2014adam}, we set $\beta_1$, $\beta_2$, and $\epsilon$ to $0.9$, $0.999$, and $10^{-8}$, respectively. As Adam can be roughly viewed as a union of SGD with Momentum and RMSProp, we may expect that if one of the two is particularly susceptible to catastrophic forgetting, so too would Adam. Thus, there is some understanding we can gain by aligning their hyperparameters with some of the hyperparameters used by Adam. So to be consistent with Adam, in RMSProp, we set the coefficient for the moving average to $0.999$ and $\epsilon$ to $10^{-8}$, and, in SGD with Momentum, we set the momentum parameter to $0.9$. In the MNIST and Fashion MNIST testbeds, we select one $\alpha$ for each of the above optimizers by trying each of $2^{-3}$, $2^{-4}$, ..., $2^{-18}$ and selecting whatever minimized the average number of steps needed to complete the four phases. As the Mountain Car testbed and Acrobot testbed are likely to be harder for the ANN to learn, we select one $\alpha$ for each of these testbeds by trying each of $2^{-3}$, $2^{-3.5}$, ..., $2^{-18}$ and selecting whatever minimized the average area under the curve of the post-episode mean squared value error. We provide a sensitivity analysis for our selection of the coefficient for the moving average in RMSProp, for the momentum parameter in SGD with Momentum, as well as for our selection of $\alpha$ with each of the four optimizers. We limit this sensitivity analysis to the retention and relearning metrics in the MNIST testbed. We extend this sensitivity analysis to the other metrics and testbeds in Appendix~\ref{app:additional_hyperparameter_sensitivity_analysis} of our supplementary material.

\section{Results}
\label{sec:results}

Since we are only interested in the phenomenon of catastrophic forgetting itself, we only report the learning systems' performance in terms of the metrics described in Section~\ref{sec:measuring_catastrophic_forgetting} here and skip reporting their performance on the actual problems. The curious reader can refer to Appendix~\ref{app:actual_performance_on_the_testbeds} of our supplementary material for that information.

\begin{figure}
    \begin{minipage}{0.48\textwidth}
        \centering
        \includegraphics[scale=0.65]{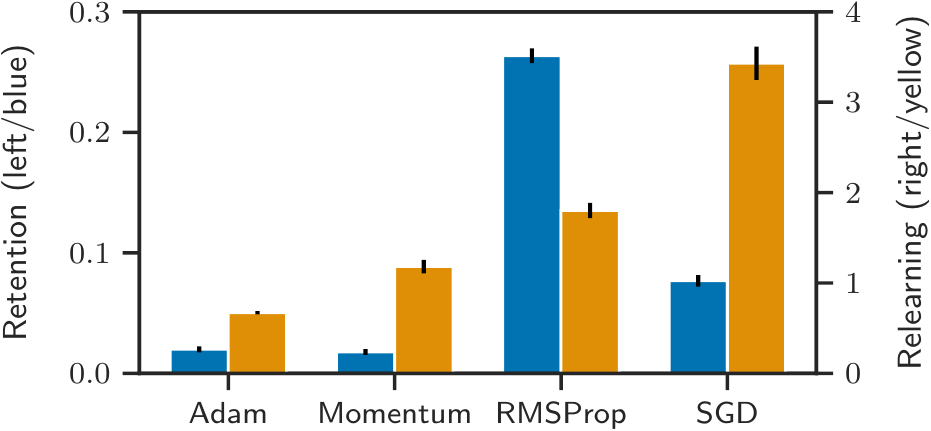}
    \end{minipage}
    \hfill
    \begin{minipage}{0.48\textwidth}
        \centering
        \includegraphics[scale=0.65]{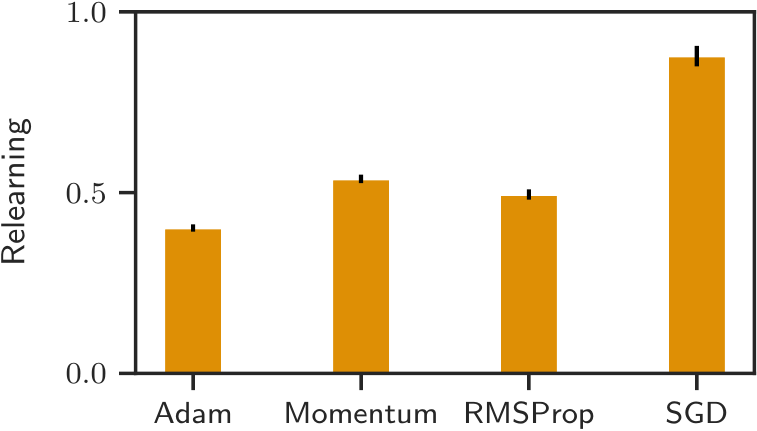}
    \end{minipage}
    \caption{Retention and relearning in the (left) MNIST testbed, and relearning in the (right) Fashion MNIST testbed (higher is better in both). Here, retention is defined as the learning system's accuracy on the first task after training it on the first task to mastery, then training it on the second task to mastery, and relearning is defined as the length of the first phase as a function of the third.}
    \label{fig:mnist_and_fashion_mnist_interference}
\end{figure}

\begin{figure}
    \begin{minipage}{0.48\textwidth}
        \centering
        \includegraphics[scale=0.575]{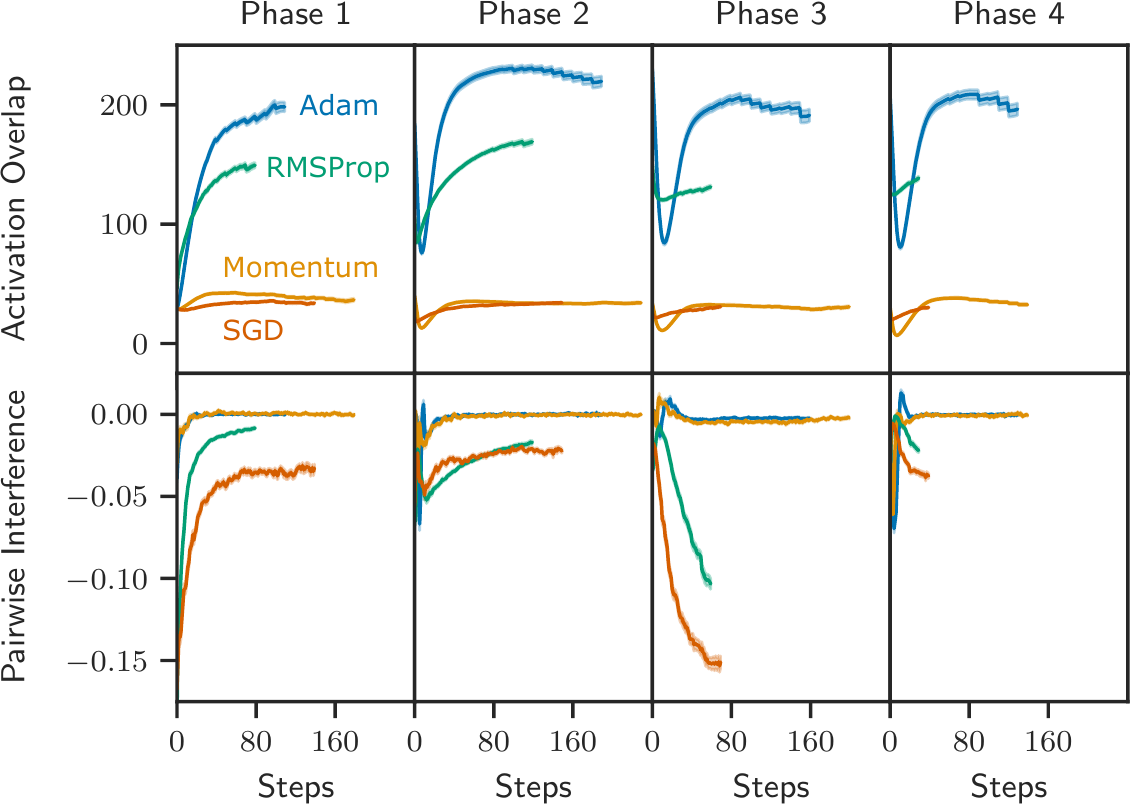}
    \end{minipage}
    \hfill
    \begin{minipage}{0.48\textwidth}
        \centering
        \includegraphics[scale=0.575]{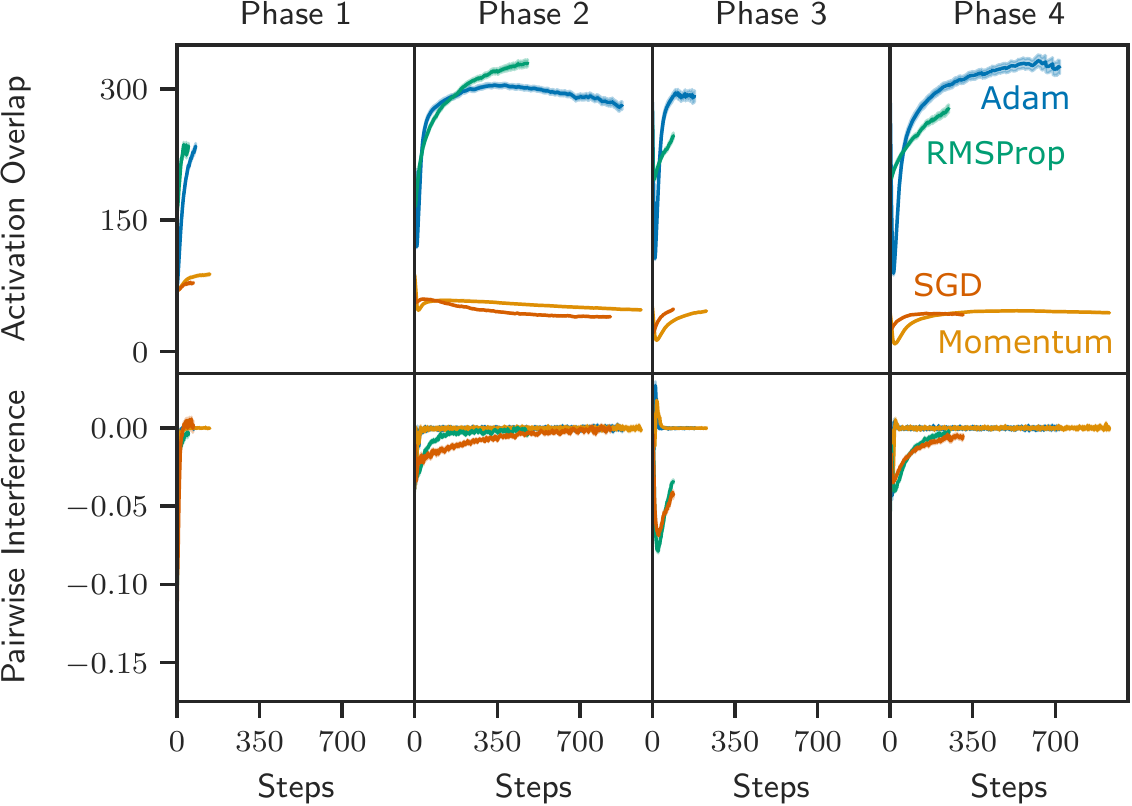}
    \end{minipage}
    \caption{Activation overlap and pairwise interference exhibited by the four optimizers as a function of phase and step in phase in the (left) MNIST testbed, and (right) Fashion MNIST testbed (lower is better in both). Lines are averages of all runs currently in that phase and are only plotted for steps where at least half of the runs for a given optimizer are still in that phase. Standard error is shown with shading but is very small.}
    \label{fig:mnist_and_fashion_mnist_additional_interference}
\end{figure}

The left side of Figure~\ref{fig:mnist_and_fashion_mnist_interference} shows the retention and relearning of the four optimizers in the MNIST testbed, and the right side shows the retention of the four optimizers in the Fashion MNIST testbed. Recall that, here, retention is defined as the learning system's accuracy on the first task after training it on the first task to mastery, then training it on the second task to mastery, and relearning is defined as the length of the first phase as a function of the third. When comparing the retention displayed by the optimizers in the MNIST testbed, RMSProp vastly outperformed the other three here. However, when comparing relearning instead, SGD is the clear leader. In the Fashion MNIST testbed, retention was less than $0.001$ with each of the four optimizers. Nevertheless, the same trend with regards to relearning in the MNIST testbed results can be observed in the Fashion MNIST testbed results.  Also notable here, Adam displayed particularly poor performance in all cases.

The left side of Figure~\ref{fig:mnist_and_fashion_mnist_additional_interference} shows the activation overlap and pairwise interference of the four optimizers in the MNIST testbed, and the right side shows these in the Fashion MNIST testbed. Note that, in Figure~\ref{fig:mnist_and_fashion_mnist_additional_interference}, lines stop when at least half of the runs for a given optimizer have moved to the next phase. Also, note that activation overlap should be expected to increase here as training progress since the network's representation for samples starts as random noise. Interestingly, the results under the MNIST and Fashion MNIST testbeds here are similar. Consistent with the retention and relearning metric, Adam exhibited the highest amount of activation overlap here. However, in contrast to the retention and relearning metric, RMSProp exhibited the second highest. Only minimal amounts are displayed with both SGD and SGD with Momentum. When compared with activation overlap, the pairwise interference reported in Figure~\ref{fig:mnist_and_fashion_mnist_additional_interference} seems to agree much more here with the retention and relearning metrics: SGD displays less pairwise interference than RMSProp, which, in turn, displays much less than either Adam or SGD with Momentum.

\begin{figure}
    \centering
    \includegraphics[scale=0.575]{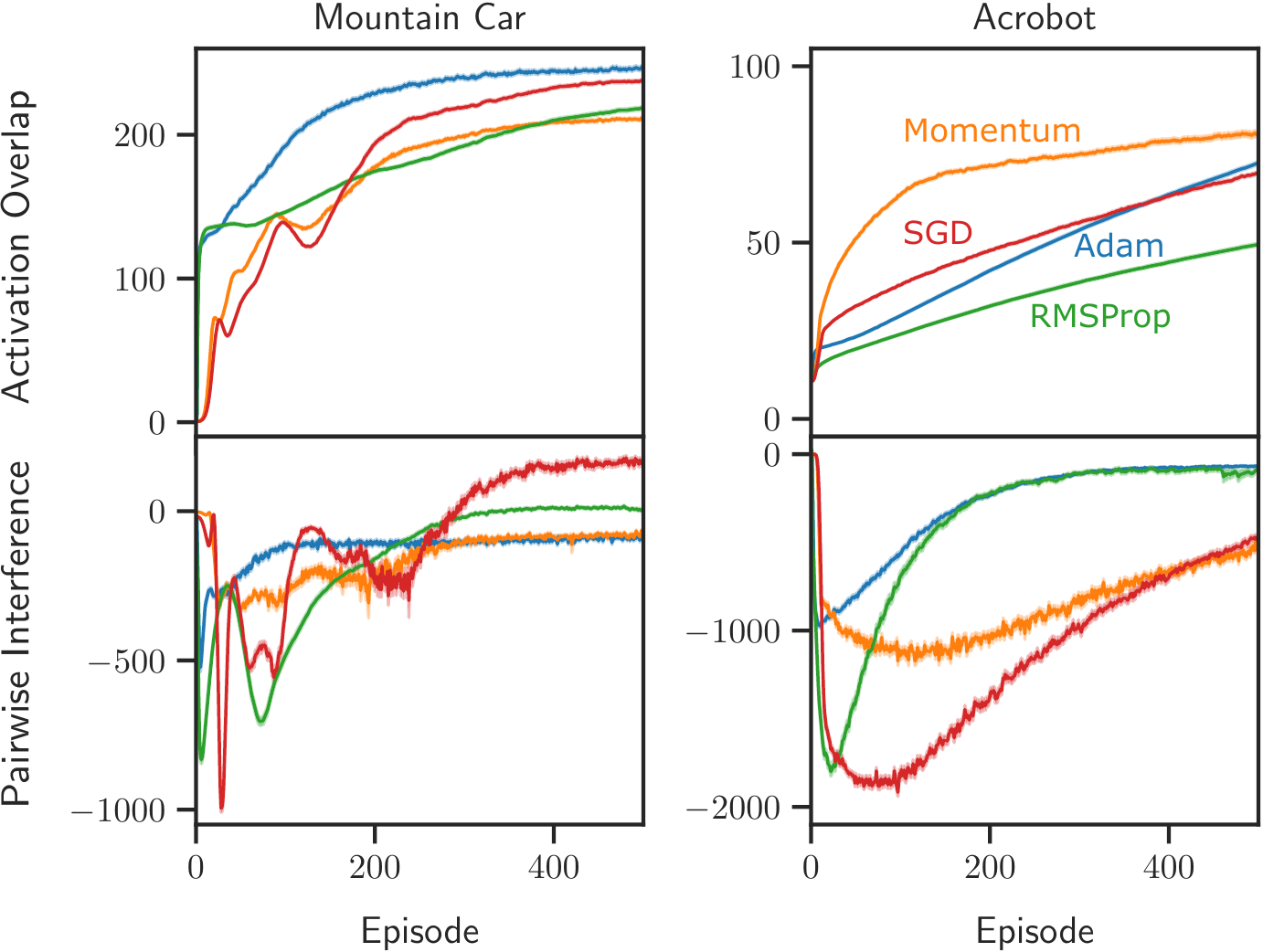}
    \caption{Activation overlap and pairwise interference exhibited by the four optimizers as a function of episode in the Mountain Car and Acrobot testbeds (lower is better). Lines are averages of all runs, and standard error is shown with shading but is very small.}
    \label{fig:mountain_car_and_acrobot_interference}
\end{figure}

Figure~\ref{fig:mountain_car_and_acrobot_interference} shows the activation overlap and pairwise interference of each of the four optimizers in the Mountain Car and Acrobot testbeds at the end of each episode. In Mountain Car, Adam exhibited both the highest mean and final activation overlap, whereas SGD with Momentum exhibited the least. However, in Acrobot, SGD with Momentum exhibited both the highest mean and final activation overlap.

When looking at the post-episode pairwise interference values shown in Figure~\ref{fig:mountain_car_and_acrobot_interference}, again, some disagreement is observed. While SGD with Momentum seemed to do well in both Mountain Car and Acrobot, vanilla SGD did well only in Acrobot and did the worst in Mountain Car. Notably, pairwise interference in Mountain Car is the only instance under any of the metrics or testbeds of Adam being among the better two optimizers.

\begin{figure}
    \begin{minipage}{0.48\textwidth}
        \centering
        \includegraphics[scale=0.625]{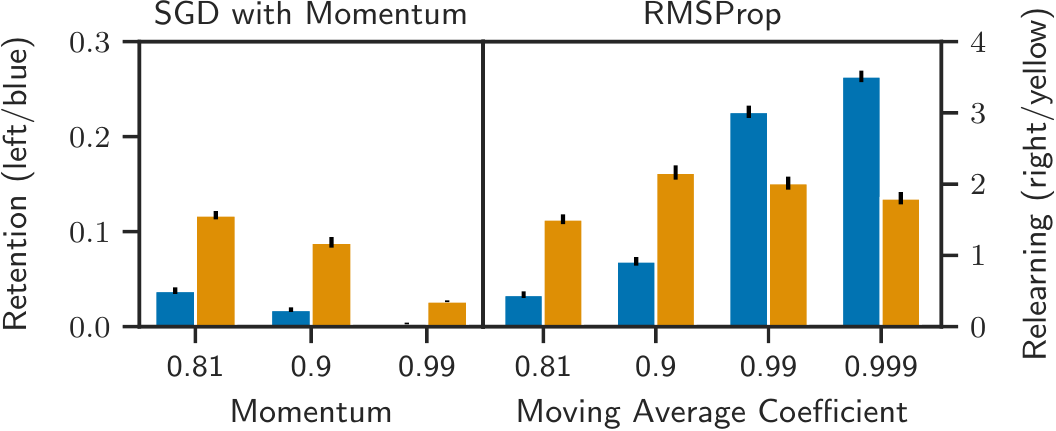}
        \caption{Retention and relearning in the MNIST testbed for SGD with Momentum under different values of momentum, and RMSProp under different coefficients for the moving average (higher is better). Other hyperparameters were set to be consistent with Figure~\ref{fig:mnist_and_fashion_mnist_interference}.}
        \label{fig:mnist_interference_momentum_and_rms}
    \end{minipage}
    \hfill
    \begin{minipage}{0.48\textwidth}
        \centering
        \includegraphics[scale=0.625]{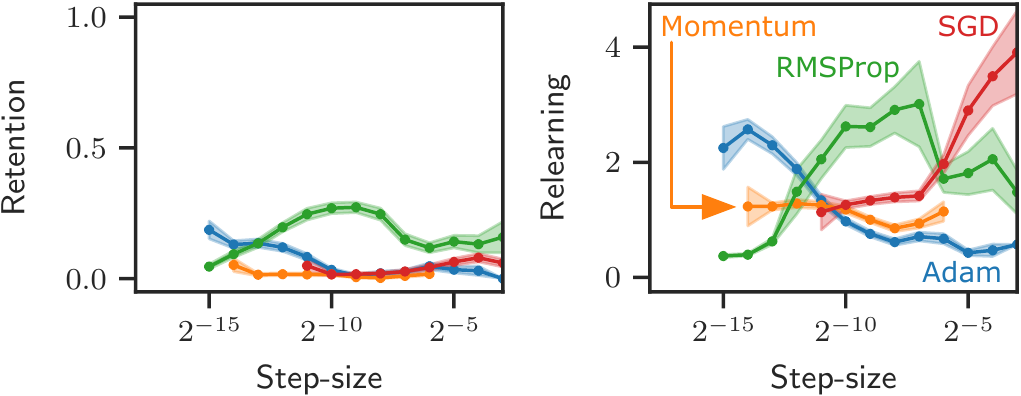}
        \caption{Retention and relearning in the MNIST testbed for each optimizer under different values of $\alpha$ (higher is better). Other hyperparameters were set to be consistent with Figure~\ref{fig:mnist_and_fashion_mnist_interference}. Lines are averages of all runs, and standard error is shown with shading. Lines are only drawn for values of $\alpha$ in which no run under the optimizer resulted in numerical instability.}
        \label{fig:mnist_interference_step-size}
    \end{minipage}
\end{figure}

Figure~\ref{fig:mnist_interference_momentum_and_rms} shows the retention and relearning in the MNIST testbed for SGD with Momentum as a function of momentum, and RMSProp as a function of the coefficient of the moving average. As would be expected with the results on SGD, lower values of momentum produce less forgetting. Conversely, lower coefficients produce worse retention in RMSProp, but seem to have less effect on relearning. Note that, under all the variations shown here, in no instance does SGD with Momentum or RMSProp outperform vanilla SGD with respect to relearning.

Similar to Figure~\ref{fig:mnist_interference_momentum_and_rms}, Figure~\ref{fig:mnist_interference_step-size} shows the retention and relearning of the four optimizers as a function of $\alpha$. While---unsurprisingly---$\alpha$ has a large effect on both metrics, the effect is smooth with similar values of $\alpha$ producing similar values for retention and relearning.

\section{Discussion}
\label{sec:discussion}

The results provided in Section~\ref{sec:results} allow us to reach several conclusions. First and foremost, as we observed a number of differences between the different optimizers over a variety of metrics and in a variety of testbeds, we can safely conclude that there can be no doubt that the choice of which modern gradient-based optimization algorithm is used to train an ANN has a meaningful and large effect on catastrophic forgetting. As we explored the most prominent of these, it is safe to conclude that this effect is likely impacting a large amount of contemporary work in the area.

\begin{figure}
    \centering
    \includegraphics[scale=0.9]{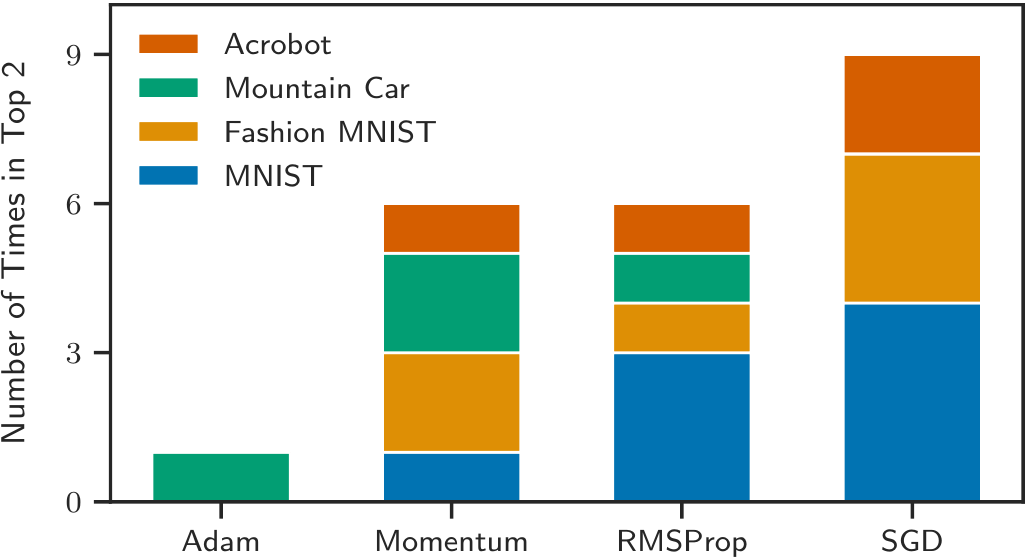}
    \caption{Number of times each optimizer ranked either first or second under a metric in a testbed. In most of our results, a natural grouping was present between a pair of optimizers that did well and a pair of optimizers that did poorly. This grouping aligns with our earlier conjecture that Adam may be particularly susceptible to catastrophic forgetting when either SGD with Momentum or RMSProp is particularly susceptible. Thus, this figure summarizes the performance of each of the optimizers over the metrics and testbeds looked at.}
    \label{fig:top_2}
\end{figure}

We earlier postulated that Adam being viewable as a combination of SGD with Momentum and RMSProp could mean that if either of the two mechanisms exacerbated catastrophic forgetting, then this would carry over to Adam. Thus, it makes sense to look at how often each of the four optimizers was either the best or second-best under a given metric and testbed. This strategy for interpreting the above results is supported by the fact that---in many of our experiments--the four optimizers could be divided naturally into one pair that did well and one pair that did poorly. The results of this process are shown in Figure~\ref{fig:top_2}. Looking at Figure~\ref{fig:top_2}, it is very obvious that Adam was particularly vulnerable to catastrophic forgetting and that SGD outperformed the other optimizers overall. These results suggest that Adam should generally be avoided---and ideally replaced by SGD---when dealing with a problem where catastrophic forgetting is likely to occur. We provide the exact rankings of the algorithms under each metric and testbed in Appendix~\ref{app:detailed_rankings}.

When looking at SGD with Momentum as a function of momentum and RMSProp as a function of the coefficient of the moving average, we saw evidence that these hyperparameters have a pronounced effect on the amount of catastrophic forgetting. Since the differences observed between vanilla SGD and SGD with Momentum can be attributed to the mechanism controlled by the momentum hyperparameter, and since the differences between vanilla SGD and RMSProp can be similarly attributed to the mechanism controlled by the moving average coefficient hyperparameter, this is in no way surprising. However, as with what we observed with $\alpha$, the relationship between the hyperparameters and the amount of catastrophic forgetting was generally smooth; similar values of the hyperparameter produced similar amounts of catastrophic forgetting. Furthermore, the optimizer seemed to play a more substantial effect here. For example, the best retention and relearning scores for SGD with Momentum we observed were still only roughly as good as the worst such scores for RMSProp. Thus while these hyperparameters have a clear effect on the amount of catastrophic forgetting, it seems unlikely that a large difference in catastrophic forgetting can be easily attributed to a small difference in these hyperparameters.

One metric that we explored was activation overlap. While \citet{french1991using} argued that more activation overlap is the cause of catastrophic forgetting and so can serve as a viable metric for it~(p.~173), in the MNIST testbed, activation overlap seemed to be in opposition to the well-established retention and relearning metrics. These results suggested that, while Adam suffers a lot from catastrophic forgetting, so too does RMSProp. Together, this suggests that catastrophic forgetting cannot be a consequence of activation overlap alone. Further studies must be conducted to understand why the unique representation learned by RMSProp here leads to it performing well on the retention and relearning metrics despite having a greater representational overlap.

On the consistency of the results, the variety of rankings we observed in Section~\ref{sec:results} validate previous concerns regarding the challenge of measuring catastrophic forgetting. Between testbeds, as well as between different metrics in a single testbed, vastly different rankings were produced. While each testbed and metric was meaningful and thoughtfully selected, little agreement appeared between them. Thus, we can conclude that, as we hypothesized, catastrophic forgetting is a subtle phenomenon that cannot be characterized by only limited metrics or limited problems.

When looking at the different metrics, the disagreement between retention and relearning is perhaps the most concerning. Both are derived from principled, crucial metrics for forgetting in psychology. As such, when in a situation where using many metrics is not feasible, we recommend ensuring that at least retention and relearning-based metrics are present. If these metrics are not available due to the nature of the testbed, we recommend using pairwise interference as it tended to agree more closely with retention and relearning than activation overlap. That being said, more work should be conducted to validate these recommendations.

\section{Conclusion}
\label{sec:conclusion}

In this work, we sought to improve our understanding of catastrophic forgetting in ANNs by revisiting the fundamental questions of (1)~how we can quantify catastrophic forgetting, and (2)~how do the choices we make when designing learning systems affect the amount of catastrophic forgetting that occurs during training. To answer these questions we explored four metrics for measuring catastrophic forgetting: retention, relearning, activation overlap, and pairwise interference. We applied these four metrics to four testbeds from the reinforcement learning and supervised learning literature and showed that (1)~catastrophic forgetting is not a phenomenon which can be effectively described by either a single metric or a single family of metrics, and (2)~the choice of which modern gradient-based optimizer is used to train an ANN has a serious effect on the amount of catastrophic forgetting.

Our results suggest that users should be wary of the optimization algorithm they use with their ANN in problems susceptible to catastrophic forgetting---especially when using Adam but less so when using SGD. When in doubt, we recommend simply using SGD without any kind of momentum and would advise against using Adam.

Our results also suggest that, when studying catastrophic forgetting, it is important to consider many different metrics. We recommend using at least a retention-based metric and a relearning-based metric. If the testbed prohibits using those metrics, we recommend using pairwise interference. Regardless of the metric used, though, research into catastrophic forgetting---like much research in AI---must be cognisant that different testbeds are likely to favor different algorithms, and results on single testbeds are at high risk of not generalizing.

\section{Future Work}
\label{sec:future_work}

While we used various testbeds and metrics to quantify catastrophic forgetting, we only applied it to answer whether one particular set of mechanisms affected catastrophic forgetting. Moreover, no attempt was made to use the testbed to examine the effect of mechanisms specifically designed to mitigate catastrophic forgetting. The decision to not focus on such methods was made as \citet{kemker2018measuring} already showed that these mechanisms' effectiveness varies substantially as both the testbed changes and the metric used to quantify catastrophic forgetting changes. \citeauthor{kemker2018measuring}, however, only considered the retention metric in their work, so some value exists in looking at these methods again under the broader set of metrics we explore here.

In this work, we only considered shallow ANNs. Contemporary deep learning frequently utilizes networks with many---sometimes hundreds---of hidden layers. While, \citet{ghiassian2020improving} showed that this might not be the most impactful factor in catastrophic forgetting~(p.~444), how deeper networks affect the nature of catastrophic forgetting remains largely unexplored. Thus further research into this is required.

One final opportunity for future research lies in the fact that, while we explored several testbeds and multiple metrics for quantifying catastrophic forgetting, there are many other, more complicated testbeds, as well as several still-unexplored metrics which also quantify catastrophic forgetting (e.g., \citet{fedus2020catastrophic}). Whether the results of this work extend to significantly more complicated testbeds remains an important open question, as is the question of whether or not these results carry over to the control case of the reinforcement learning problem. Notably, though, it remains an open problem how exactly forgetting should be measured in the control case.

\section*{Acknowledgements}

The authors would like to thank Patrick Pilarsky and Mark Ring for their comments on an earlier version of this work. The authors would also like to thank Compute Canada for generously providing the computational resources needed to carry out the experiments contained herein. This work was partially funded by the European Research Council Advanced Grant AlgoRNN to J{\"{u}}rgen Schmidhuber (ERC no: 742870).

\nocite{brockman2016openai}
\nocite{geramifard2015rlpy}
\nocite{ghiassian2017first}
\nocite{glorot2010understanding}
\nocite{glorot2011deep}
\nocite{he2015delving}
\nocite{jarrett2009best}
\nocite{nair2010rectified}

\printbibliography

\clearpage
\appendix

\section{Additional Experimental Setup}
\label{app:additional_experimental_setup}

\paragraph{MNIST}

\begin{figure}[h]
    \centering
    \includegraphics[width=0.75\textwidth]{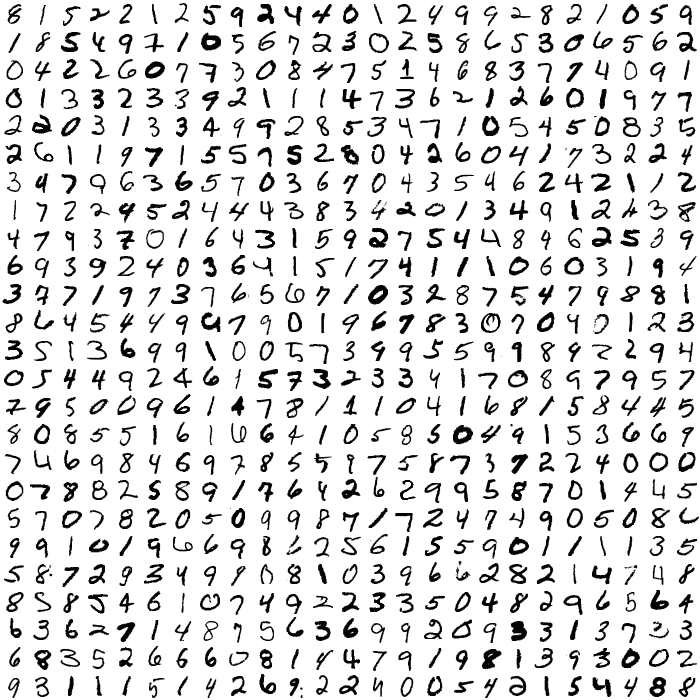}
    \caption{Some of the handwritten digits as they appear in the full MNIST dataset. Each digit appears in the dataset as a labelled \(28 \times 28\) greyscale image.}
    \label{fig:mnist_sample}
\end{figure}

The MNIST dataset contains \(28 \times 28\) greyscale images labelled according to what digit the writer was trying to inscribe. A sample of some of the images comprising the MNIST dataset is shown in Figure~\ref{fig:mnist_sample}. For the MNIST testbed, we use only the ones, twos, threes, and fours from the MNIST dataset to generate two independent two-class classification tasks. In each of the two tasks, the learning system must predict which of the four digits a given image corresponds to. While the network is only solving one of the two tasks at a given timestep, no information is explicitly provided to the network that would allow it to discern when the task it is solving changes. Thus it is free to guess that, for example, a given image is a four when the task it is currently solving contains only images of ones and twos.

To build the data-stream for the MNIST testbed, we use stratified random sampling to divide the MNIST dataset into ten folds of approximately $6000$ examples each. The exact distribution of classes in the folds is provided in Appendix~\ref{app:distribution_of_digits_in_mnist_and_fashion_mnist_folds}. We use two folds to select hyperparameters for the learning systems and two folds to evaluate the learning systems under these hyperparameters. To prevent the data-stream from ever presenting the same example to the learning system more than once, we always used one fold for the first two phases and one fold for later phases. To create a dataset we can use to obtain a setting-wide measure of activation overlap and pairwise interference, we sample ten examples from each of the four classes out of the unused folds.

\paragraph{Fashion MNIST}

\begin{figure}[h]
    \centering
    \includegraphics[width=0.75\textwidth]{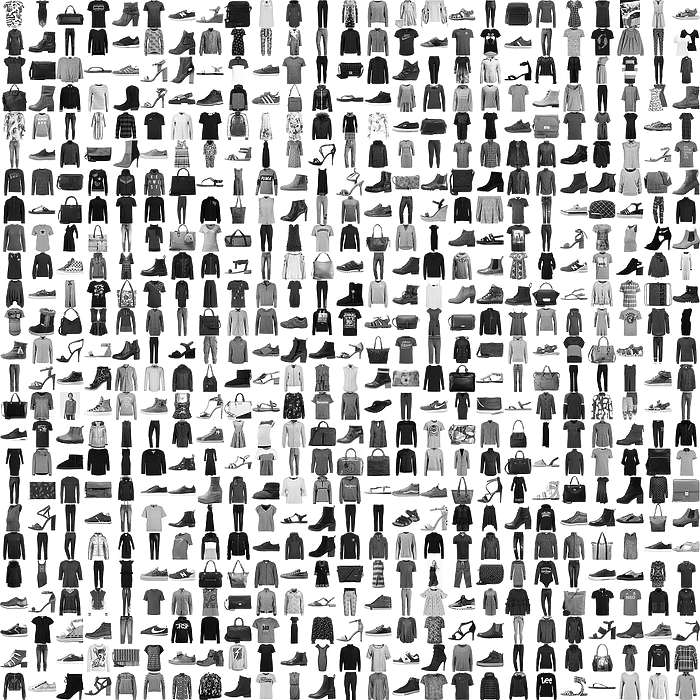}
    \caption{Some of the pictures of clothes as they appear in the full Fashion MNIST dataset. Each picture appears in the dataset as a labelled \(28 \times 28\) greyscale image.}
    \label{fig:fashion_mnist_sample}
\end{figure}

The Fashion MNIST dataset was designed to be nearly identical to the MNIST dataset, but instead of handwritten digits, it uses pictures of clothes. The different classes then correspond to different types of clothing rather than different digits. A sample of some of the images comprising the Fashion MNIST dataset is shown in Figure~\ref{fig:fashion_mnist_sample}. We exploit the similarity of MNIST and Fashion MNIST and follow the exact same procedure to derive a testbed from the Fashion MNIST dataset.

\paragraph{Mountain Car}

\begin{figure}[h]
    \centering
    \includegraphics[width=0.7\textwidth]{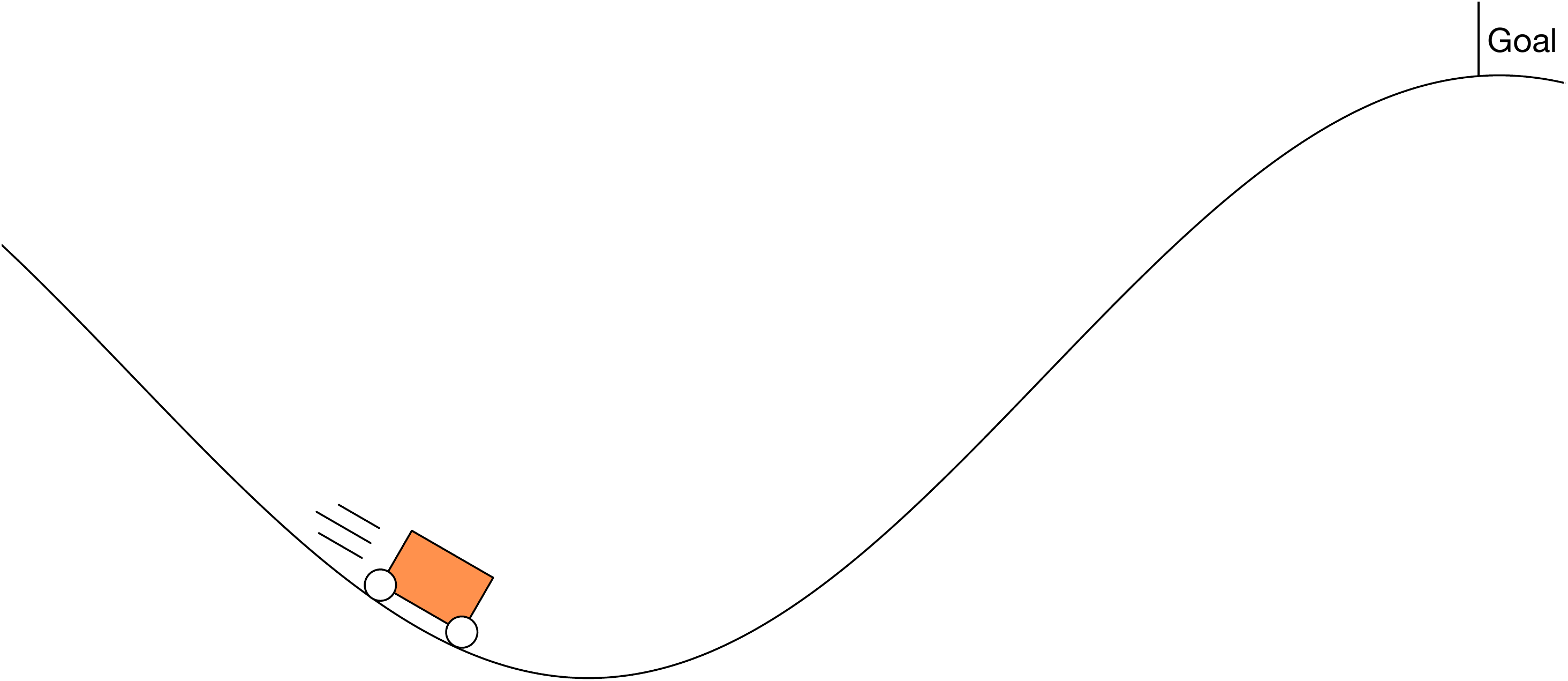}
    \caption{The Mountain Car testbed simulates a car (shown in orange) whose objective is to reach the goal on the right. The car starts at the bottom of the valley and must rock back and forth in order to climb the mountain. Note that the car is prevented from falling off the left edge of the world by an invisible wall.}
    \label{fig:mountain_car_diagram}
\end{figure}

Our Mountain Car testbed is based on the popular classic reinforcement learning domain that models a car trying to climb a hill (see Figure~\ref{fig:mountain_car_diagram}). The car starts at the bottom of a valley and lacks sufficient power to make it up the mountain by acceleration alone. Instead, it must rock back and forth to build up enough momentum to climb the mountain.

Formally, Mountain Car is an undiscounted episodic domain where, at each step, the car measures its position \(p \in [-1.2, 0.6]\) and velocity \(v \in [-0.07, 0.07]\), and then either accelerates in the direction of the goal, decelerate, or does neither. To capture the idea that the car should reach the goal quickly, it receives a reward of $-1$ at each step. Each episode begins with $v = 0$ and $p$ selected uniformly from $[-0.6, 0.4)$, and ends when \(p \geq 0.5\). If, at any point, \(p \leq -1.2\), then $p$ is set to be equal to $-1.2$ and $v$ is set to be equal to $0$. This last rule simulates the effect of it harmlessly hitting an impassable wall. With this last rule in mind, the position and velocity of the car in Mountain Car is updated at each step according to the following equations:
\begin{align*}
    p_{t + 1} &= p_{t} + v_{t + 1} \\
    v_{t + 1} &= v_{t} + 0.001 a_t - 0.0025 \text{cos}(3 p_{t})
\end{align*}
where $a_{t} = 0$ when decelerating, $a_{t} = 2$ when accelerating, and $a_{t} = 1$ when the action selected is to do neither.

\begin{figure}
    \centering
    \includegraphics[width=0.6\textwidth]{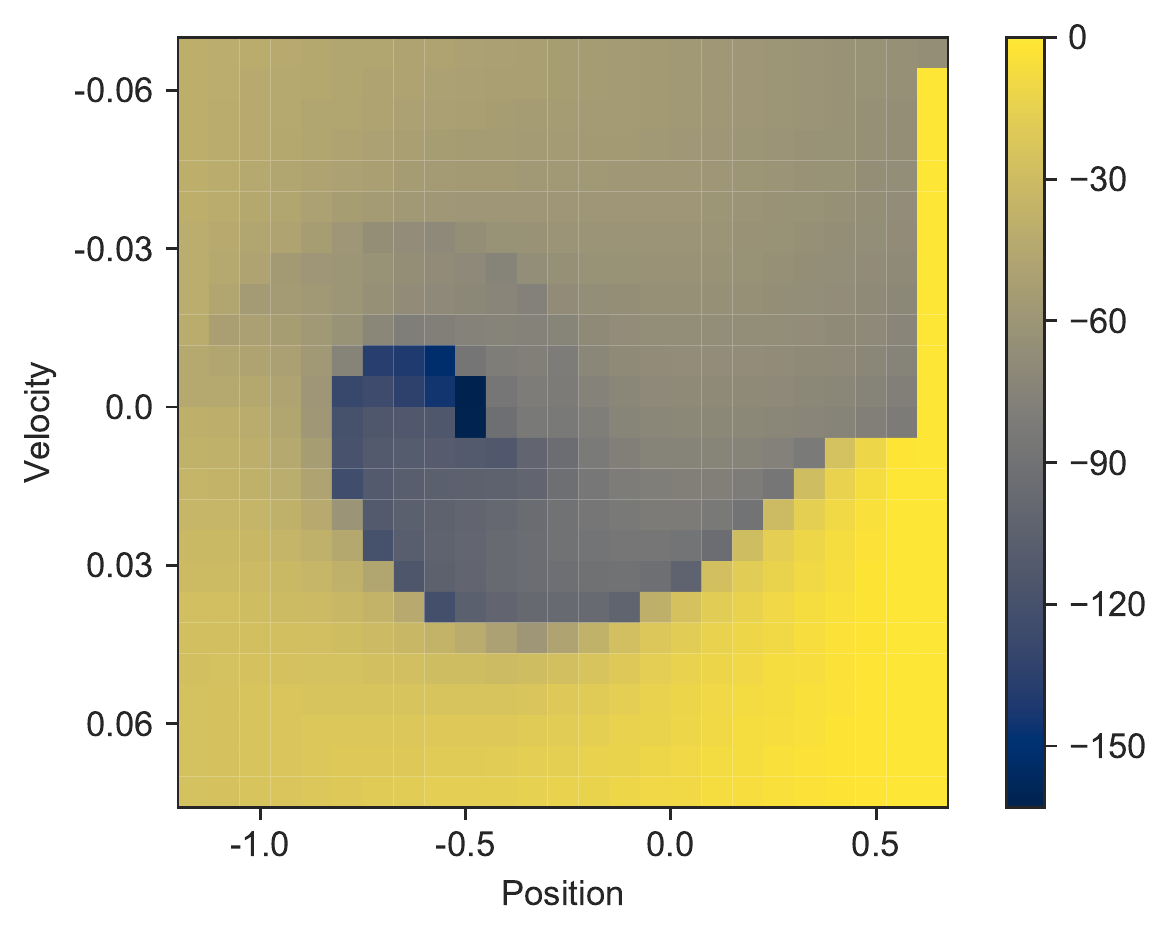}
    \caption{Values of states in Mountain Car domain when the policy the car follows is to always accelerate in the direction of movement. Note that the value of a state in Mountain Car is the negation of the expected number of steps before the car reaches the goal.}
    \label{fig:mountain_car_state_values}
\end{figure}

In our testbed, we use a fixed policy where---as in \citet{ghiassian2017first}---the agent always accelerate in the direction of motion or, if it is stationary, does not accelerate at all. We plot the state-values in Mountain Car under the above policy in Figure~\ref{fig:mountain_car_state_values}. The learning system's goal in this testbed is to learn, for each timestep, what the value of the current state is. In Mountain Car, this value corresponds to the expected number of steps left in the episode.

To measure performance in this testbed, we look at the Root Mean Squared Value Error, or RMSVE under the above policy which is defined to be
\begin{equation*}
    \text{RMSVE}(\theta) = \sqrt{\sum\limits_{s \in \mathcal{S}} d_{\pi}(s) (\hat{v}_{\pi}(s; \theta) - v_{\pi}(s))^2}
\end{equation*}
where $\mathcal{S}$ is the set of all states, $d_{\pi}(s)$ is the proportion of time above policy $\pi$ spends in state $s$, $\hat{v}_{\pi}(s)$ is the value estimate for state $s$ under $\pi$, and $v_{\pi}(s)$ is the true value of state $s$ under $\pi$. We approximate performance here by by repeatedly running episodes to create a trajectory containing 10,000,000 transitions and then sampling $500$ states from this trajectory uniformly and with replacement. To create the dataset for measuring activation overlap and pairwise interference in Mountain Car, we overlay a \(6 \times 6\) evenly-spaced grid over the state space (with position only up to the goal position) and then using the center points of the cells in this grid as examples.

\paragraph{Acrobot}

\begin{figure}[h]
    \centering
    \includegraphics[width=0.5\columnwidth]{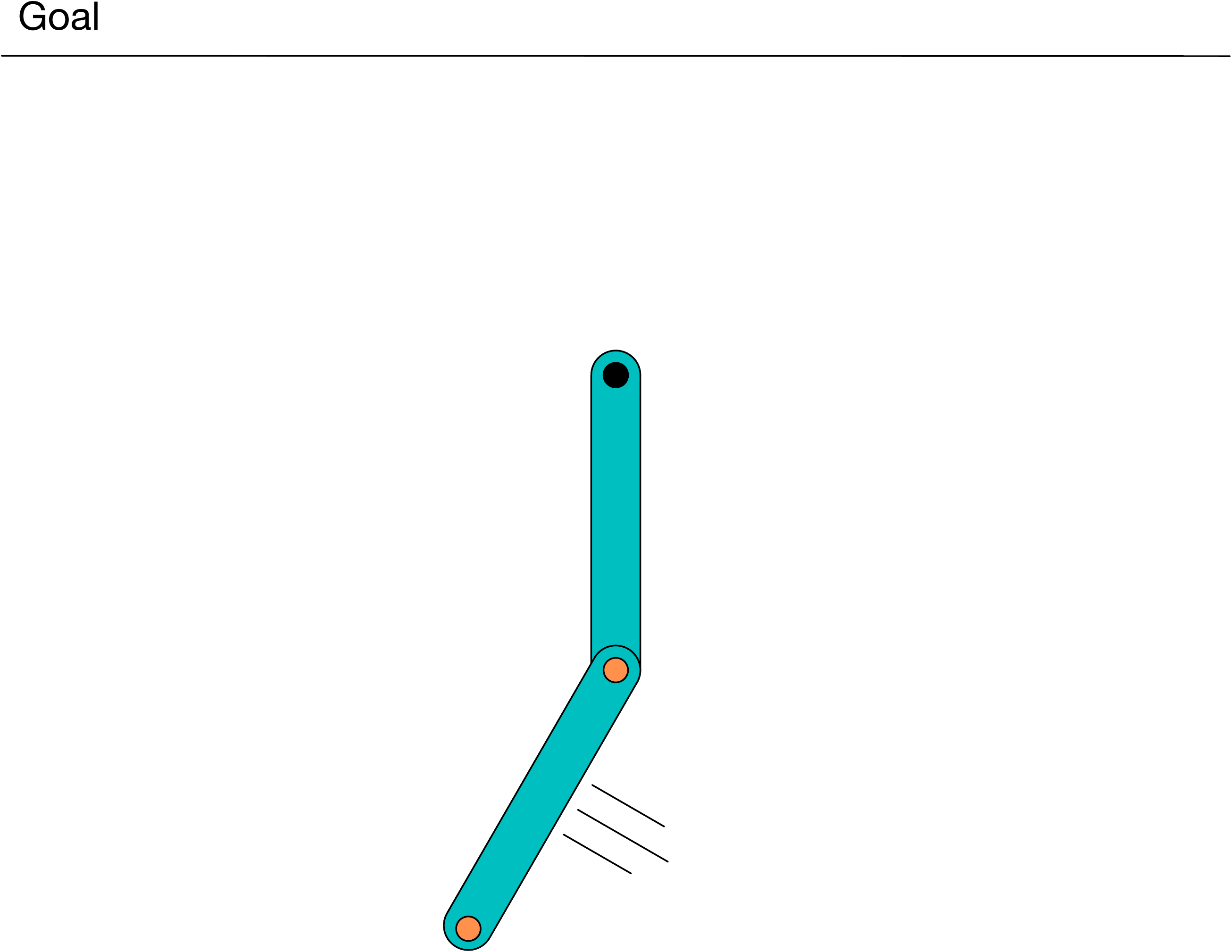}
    \caption{The Acrobot testbed simulates a double pendulum whose objective is to place the end of the outer pendulum above a goal line. Force is applied to the joint between the two pendulums. The pendulums must rock back and forth in order for the outer pendulum to reach the goal.}
    \label{fig:acrobot_diagram}
\end{figure}

Our acrobot testbed is---like Mountain Car---based on the popular, classic reinforcement learning domain. It models a double pendulum combating gravity in an attempt to invert itself (see Figure~\ref{fig:acrobot_diagram}). The pendulum moves through the application of force to the joint connecting the two pendulums. However, not enough force can be applied to smoothly push the pendulum such that it becomes inverted. Instead, like in Mountain Car, the force must be applied in such a way that the pendulums build momentum by swinging back and forth.

Formally, Acrobot is an undiscounted episodic domain where, at each step, the acrobot measures the sin and cos of the angles of both joints as well as their velocities. A fixed amount of force can then be optionally applied to the joint between the two pendulums in either direction. Like with Mountain Car, the acrobot receives a reward of $-1$ at each step. Both pendulums have equal lengths, and episodes terminate when the end of the second pendulum is at least the pendulum's length above the pivot. The velocity of the inner joint angle in radian per second is bounded by \([-4\pi, 4\pi]\), and the velocity of the outer joint angle is bounded by \([-9\pi, 9\pi]\).

We use the OpenAI Gym implementation of Acrobot~\citep{brockman2016openai}, which is based on the RLPy version~\citep{geramifard2015rlpy}. The equations of motion that describe the pendulum movements are significantly more complicated than the equations for Mountain Car, and so are omitted here. The original equations of motion for Acrobot can be found on page 1044 of \citet{sutton1995generalization}, and the implementation we use can be found at \url{https://github.com/openai/gym/blob/master/gym/envs/classic_control/acrobot.py}.

Like for Mountain Car, we fix the policy of the agent. However, finding a good, simple rule-based policy for Acrobat is not as straightforward. Inspired by the policy we used in Mountain Car, we adopt a policy whereby force is applied at each step according to the direction of motion of the inner joint. To deal with situations where centripetal force renders the inner pendulum effectively immobile, we augment this policy with the rule that no force is applied if the outer joint's velocity is at least ten times greater than the velocity of the inner joint. As with Mountain Car, the learning system's goal here is to learn, for each timestep, what the current state's value is, and---like with Mountain Car---in Acrobot, this corresponds to the expected number of steps left in the episode.

We ran the above policy for 1,000,000 episodes and observed an average episode length of $156.0191$ with a standard deviation of $23.4310$ steps. The maximum number of steps in an episode was $847$, and the minimum was $109$. Thus we believe that this policy displays sufficient consistency to be useful for learning but enough variability to ensure a reasonably heterogeneous data-stream.

For consistency, to measure performance in the Acrobot testbed, we follow the same procedure as in Mountain Car. To create the dataset for measuring activation overlap and pairwise interference in Acrobot, we sample $180$ random states uniformly from the state space.

\paragraph{Network Architectures}

For each of the four testbeds, we use a feedforward ANN trained through backpropagation~\citep{rumelhart1986learning}. For the MNIST and Fashion MNIST testbeds, we use a network with one hidden layer of $100$ units and initialize all the weights by sampling from a gaussian distribution with mean $0$ and a standard deviation of $0.1$. For the Mountain Car testbed, we follow \citet{ghiassian2020improving} in using a network with one hidden layer of $50$ units with all bias weights initialized as in the MNIST and Fashion MNIST network, and Xavier initialization~\citep{glorot2010understanding} used for all the other weights. Finally, for the Acrobot testbed, we follow \citet{liu2019sparse} in using a network with two hidden layers of $32$ then $256$ units with all bias weights initialized as in the MNIST and Fashion MNIST network, and He initialization~\citep{he2015delving} used for all the other weights. We use ReLU activation~\citep{jarrett2009best,nair2010rectified,glorot2011deep} for all of the hidden layers in each of the four testbeds. For the MNIST and Fashion MNIST testbed, we use cross-entropy as our loss function, and for Mountain Car and Acrobot, we use the squared temporal-difference error as a loss function.

\paragraph{Random Seeds}

We ran each experiment with 50 different seeds to perform the $\alpha$ selection procedure and to perform the sensitivity analysis for $\alpha$. After selecting the best $\alpha$ in each scenario, we then used it with 500 other seeds to generate the results reported in Section~\ref{sec:results}. Each seed was also used to initialize the networks and control stochasticity in the testbeds. In the MNIST and Fashion MNIST testbeds, this stochasticity manifested as the order of the examples in each fold. In the Mountain Car and Acrobot testbeds, this stochasticity appears as the agent's initial state in each episode.

\paragraph{Computational Resources Used}

All experiments were run on the Compute Canada Cedar cluster. The requested allocation consisted of 4428 hours on nodes with 6 CPU cores and 8000 megabytes of memory, and 25 hours on nodes with 6 CPU cores and 64000 megabytes of memory. Plotting was done separately on a standard laptop and did not require significant computational resources.

\section{Distribution of Digits in MNIST and Fashion MNIST Folds}
\label{app:distribution_of_digits_in_mnist_and_fashion_mnist_folds}

Both the MNIST and Fashion MNIST datasets are divided into a training set and a holdout set. We constructed our MNIST and Fashion MNIST testbeds by applying stratified random sampling to their training sets to generate ten folds for each of the datasets. Our experiments did not use the holdout sets. The distribution of digits across the resulting folds for the MNIST dataset is shown in Table~\ref{tab:mnist_folds}. For the Fashion MNIST dataset, each of the ten folds contained exactly 600 examples from each class.

\begin{table}[h]
    \caption{Distribution of digits in MNIST after dividing it into a holdout set and ten stratified folds.}
    \label{tab:mnist_folds}
    \centering
    \begin{tabular}{lcccccccccc}
        \toprule
        \diagbox[width=7em]{Fold}{Digit} & 0 & 1 & 2 & 3 & 4 & 5 & 6 & 7 & 8 & 9 \\
        \midrule
        0 & 593 & 675 & 596 & 614 & 585 & 543 & 592 & 627 & 586 & 595 \\
        1 & 593 & 675 & 596 & 613 & 585 & 542 & 592 & 627 & 585 & 595 \\
        2 & 593 & 674 & 596 & 613 & 584 & 542 & 592 & 627 & 585 & 595 \\
        3 & 592 & 674 & 596 & 613 & 584 & 542 & 592 & 627 & 585 & 595 \\
        4 & 592 & 674 & 596 & 613 & 584 & 542 & 592 & 627 & 585 & 595 \\
        5 & 592 & 674 & 596 & 613 & 584 & 542 & 592 & 626 & 585 & 595 \\
        6 & 592 & 674 & 596 & 613 & 584 & 542 & 592 & 626 & 585 & 595 \\
        7 & 592 & 674 & 596 & 613 & 584 & 542 & 592 & 626 & 585 & 595 \\
        8 & 592 & 674 & 595 & 613 & 584 & 542 & 591 & 626 & 585 & 595 \\
        9 & 592 & 674 & 595 & 613 & 584 & 542 & 591 & 626 & 585 & 594 \\
        Holdout & 980 & 1135 & 1032 & 1010 & 982 & 892 & 958 & 1028 & 974 & 1009 \\
        \bottomrule
    \end{tabular}
\end{table}

\section{Actual Performance on the Testbeds}
\label{app:actual_performance_on_the_testbeds}

In the MNIST testbed, performance is measured as the time taken to transition through all four phases. The four optimizers' performance on the MNIST testbed is shown in Table~\ref{tab:mnist_speed}. Here, RMSProp outperforms the other optimizers but is closely followed by SGD. Adam clearly performs the worst here. Rankings under retention, relearning, and pairwise interference---but not activation overlap---all correspond relatively well to this ordering.

Like with the MNIST testbed, in the Fashion MNIST testbed, performance is measured as the time taken to transition through all four phases. The four optimizers' performance on the Fashion MNIST testbed is shown in Table~\ref{tab:fashion_mnist_speed}. As with MNIST, here, RMSProp generally outperforms the other optimizers but is closely followed by SGD. Adam and SGD with Momentum perform particularly poorly here. Only rankings under pairwise interference correspond well to this ordering.

\begin{table}[h]
    \caption{Average number of steps each of the four optimizers took to complete each phase in the MNIST testbed.}
    \label{tab:mnist_speed}
    \centering
    \begin{tabular}{lcccc}
        \toprule
        Optimizer & Phase 1 & Phase 2 & Phase 3 & Phase 4 \\
        \midrule
        Adam & 82.98$\pm$1.78 & 161.58$\pm$1.80 & 136.14$\pm$1.78 & 110.78$\pm$1.45 \\
        Momentum & 135.88$\pm$2.86 & 192.18$\pm$2.38 & 155.03$\pm$2.67 & 116.55$\pm$1.90 \\
        RMSProp & {\bf 60.19$\pm$1.25} & {\bf 100.08$\pm$1.28} & {\bf 49.29$\pm$1.11} & {\bf 24.54$\pm$0.81} \\
        SGD & 105.67$\pm$2.26 & 120.82$\pm$1.97 & 52.12$\pm$1.51 & 29.81$\pm$0.90 \\
        \bottomrule
    \end{tabular}
\end{table}

\begin{table}[h]
    \caption{Average number of steps each of the four optimizers took to complete each phase in the Fashion MNIST testbed.}
    \label{tab:fashion_mnist_speed}
    \centering
    \begin{tabular}{lcccc}
        \toprule
        Optimizer & Phase 1 & Phase 2 & Phase 3 & Phase 4 \\
        \midrule
        Adam & 59.89$\pm$1.45 & 731.52$\pm$9.15 & 153.16$\pm$1.59 & 573.14$\pm$9.30 \\
        Momentum & 107.46$\pm$2.17 & 815.92$\pm$9.61 & 205.26$\pm$1.74 & 754.86$\pm$10.88 \\
        RMSProp & {\bf 36.39$\pm$0.94} & {\bf 375.22$\pm$7.84} & 80.90$\pm$1.18 & {\bf 186.18$\pm$6.33} \\
        SGD & 58.21$\pm$1.47 & 630.74$\pm$11.65 & {\bf 75.86$\pm$1.24} & 233.32$\pm$9.96 \\
        \bottomrule
    \end{tabular}
\end{table}

\begin{figure}[h!]
    \centering
    \includegraphics[width=0.575\textwidth]{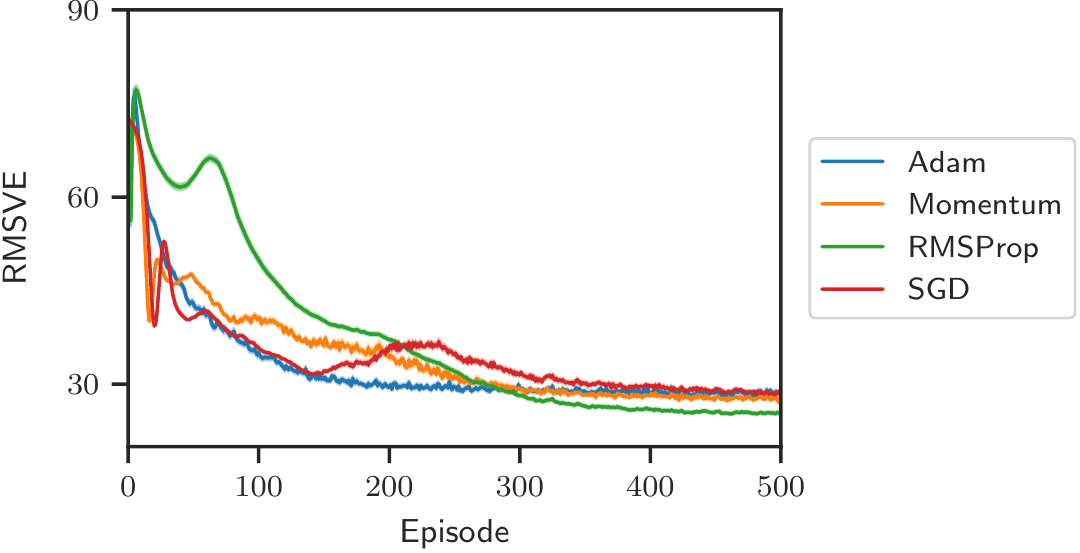}
    \caption{Performance of the four optimizers as a function of episode in the Mountain Car testbed (lower is better). Lines are averages of all runs, and standard error is shown with shading but is very small.}
    \label{fig:mountain_car_accuracies}
\end{figure}

Figure~\ref{fig:mountain_car_accuracies} shows the performance of the four optimizers in the Mountain Car testbed, as measured by RMSVE. Here, the four optimizers show relatively similar performance overall, and while RMSProp does poorly initially, it slightly outperforms the other optimizers later on.

\begin{figure}[h!]
    \centering
    \includegraphics[width=0.575\textwidth]{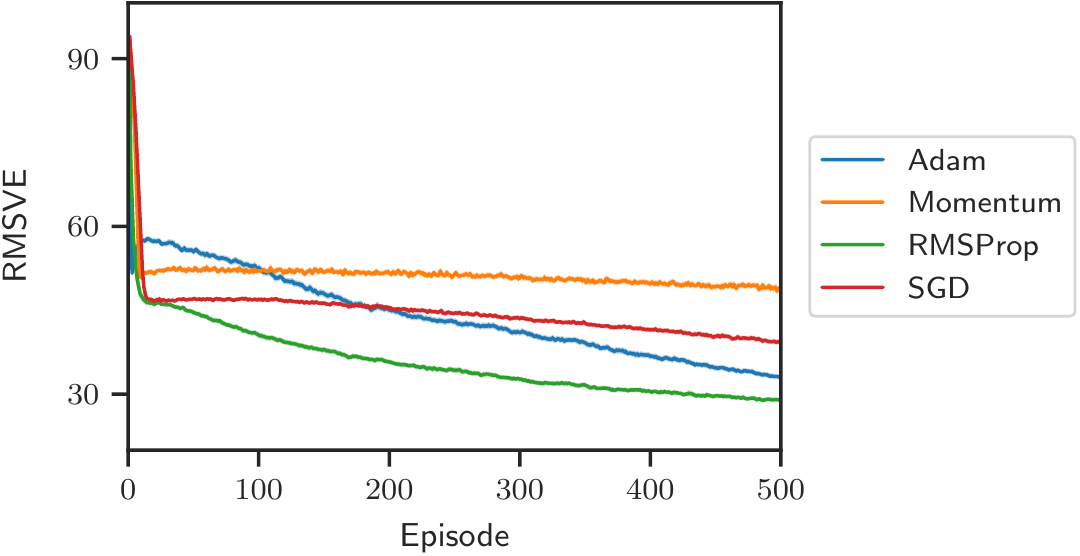}
    \caption{Performance of the four optimizers as a function of episode in the Acrobot testbed (lower is better). Lines are averages of all runs, and standard error is shown with shading but is very small.}
    \label{fig:acrobot_accuracies}
\end{figure}

Figure~\ref{fig:acrobot_accuracies} shows the performance of the four optimizers in the Acrobot testbed. Unlike in Mountain Car, the four optimizers perform at very different levels here. Unquestionably, RMSProp outperforms the other optimizers. Additionally, while Adam is slow to learn initially, it overtakes SGD and SGD with Momentum after only about $250$ episodes. These results correspond only vaguely with the ranking under activation overlap but not at all with the rankings under pairwise interference.

\section{Detailed Rankings}
\label{app:detailed_rankings}

Table~\ref{tab:rankings} ranks each of the four optimizers under the different metrics and testbeds. For Mountain Car and Acrobot, rankings under activation overlap and pairwise interference use their final values. Note that retention was almost nonexistent in the Fashion MNIST testbed with each of the four optimizers, so no meaningful ranking is possible.

\begin{table}[h]
    \caption{Rankings of optimizers under different metrics and testbeds.}
    \label{tab:rankings}
    \centering
    \begin{tabular}{llcccc}
        \toprule
        Metric & Testbed & Adam & Momentum & RMSProp & SGD \\
        \midrule
        \multirow{2}{*}{Retention} & MNIST & =3 & =3 & {\bf 1} & 2 \\
        & Fashion MNIST & - & - & - & - \\
        \midrule
        \multirow{2}{*}{Relearning} & MNIST & 4 & 3 & 2 & {\bf 1} \\
        & Fashion MNIST & 4 & 2 & 3 & {\bf 1} \\
        \midrule
        \multirow{4}{*}{Activation Overlap} & MNIST & 4 & {\bf =1} & 3 & {\bf =1} \\
        & Fashion MNIST & =3 & {\bf =1} & =3 & {\bf =1} \\
        & Mountain Car & 4 & {\bf =1} & 2 & 3 \\
        & Acrobot & 3 & 4 & {\bf =1} & 2 \\
        \midrule
        \multirow{4}{*}{Pairwise Interference} & MNIST & =3 & =3 & 2 & {\bf 1} \\
        & Fashion MNIST & =3 & =3 & {\bf =1} & {\bf =1} \\
        & Mountain Car & {\bf =1} & {\bf =1} & 3 & 4 \\
        & Acrobot & =3 & {\bf =1} & =3 & 2 \\
        \bottomrule
    \end{tabular}
\end{table}

\section{Additional Hyperparameter Sensitivity Analysis}
\label{app:additional_hyperparameter_sensitivity_analysis}

In the main text, we provided a sensitivity analysis under retention and relearning for our selection of the coefficient for the moving average in RMSProp, for the momentum parameter in SGD with Momentum, as well as for our selection of $\alpha$ with each of the four optimizers. Here we extend this sensitivity analysis to the other metrics and testbeds.

\begin{figure}
    \centering
    \includegraphics[scale=0.9]{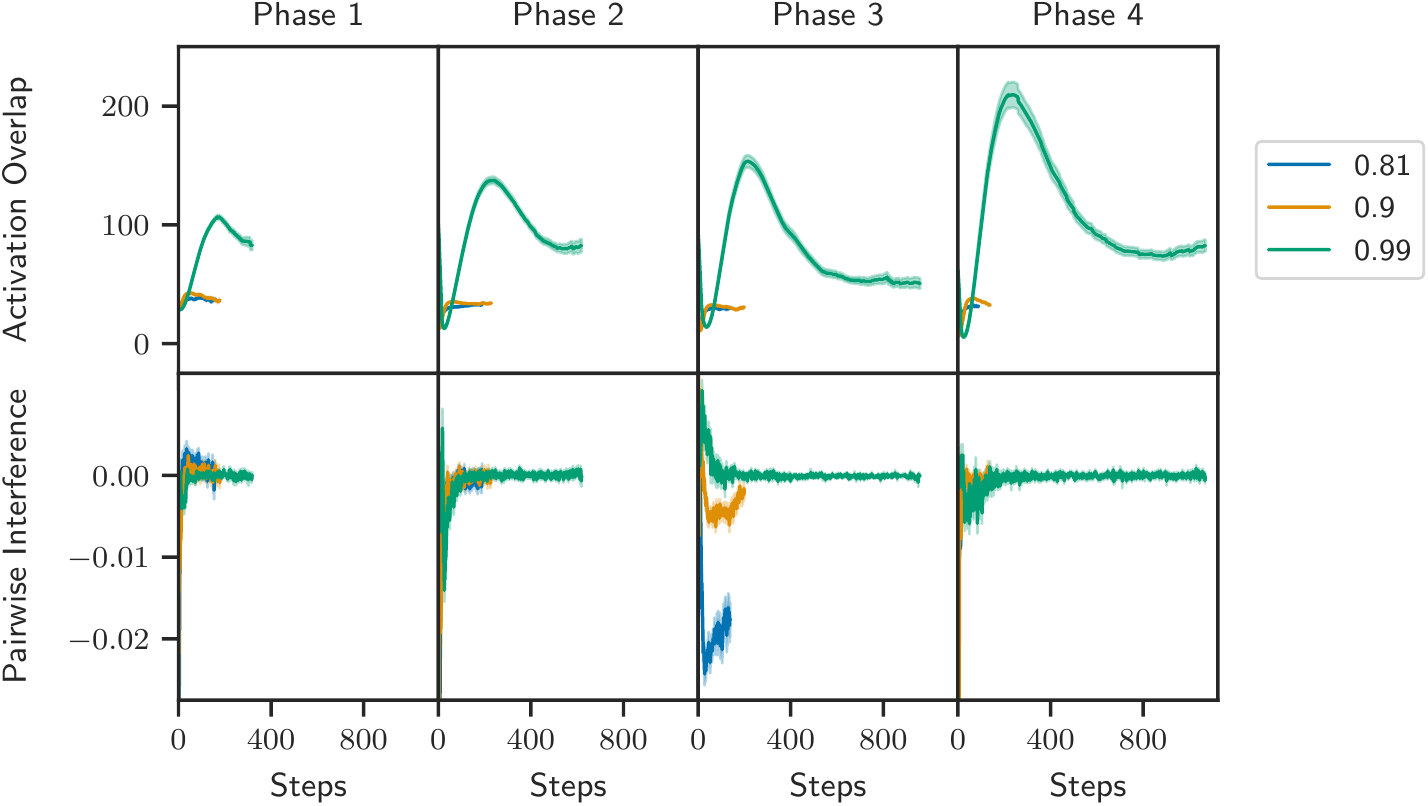}
    \caption{Activation overlap and pairwise interference in the MNIST testbed for SGD with Momentum under different values of momentum (lower is better). Other hyperparameters were set to be consistent with Figure~\ref{fig:mnist_and_fashion_mnist_additional_interference}. Lines are averages of all runs currently in that phase and are only plotted for steps where at least half of the runs for a given optimizer are still in that phase. Standard error is shown with shading but is very small.}
    \label{fig:mnist_additional_interference_momentum}
\end{figure}

\begin{figure}
    \centering
    \includegraphics[scale=0.9]{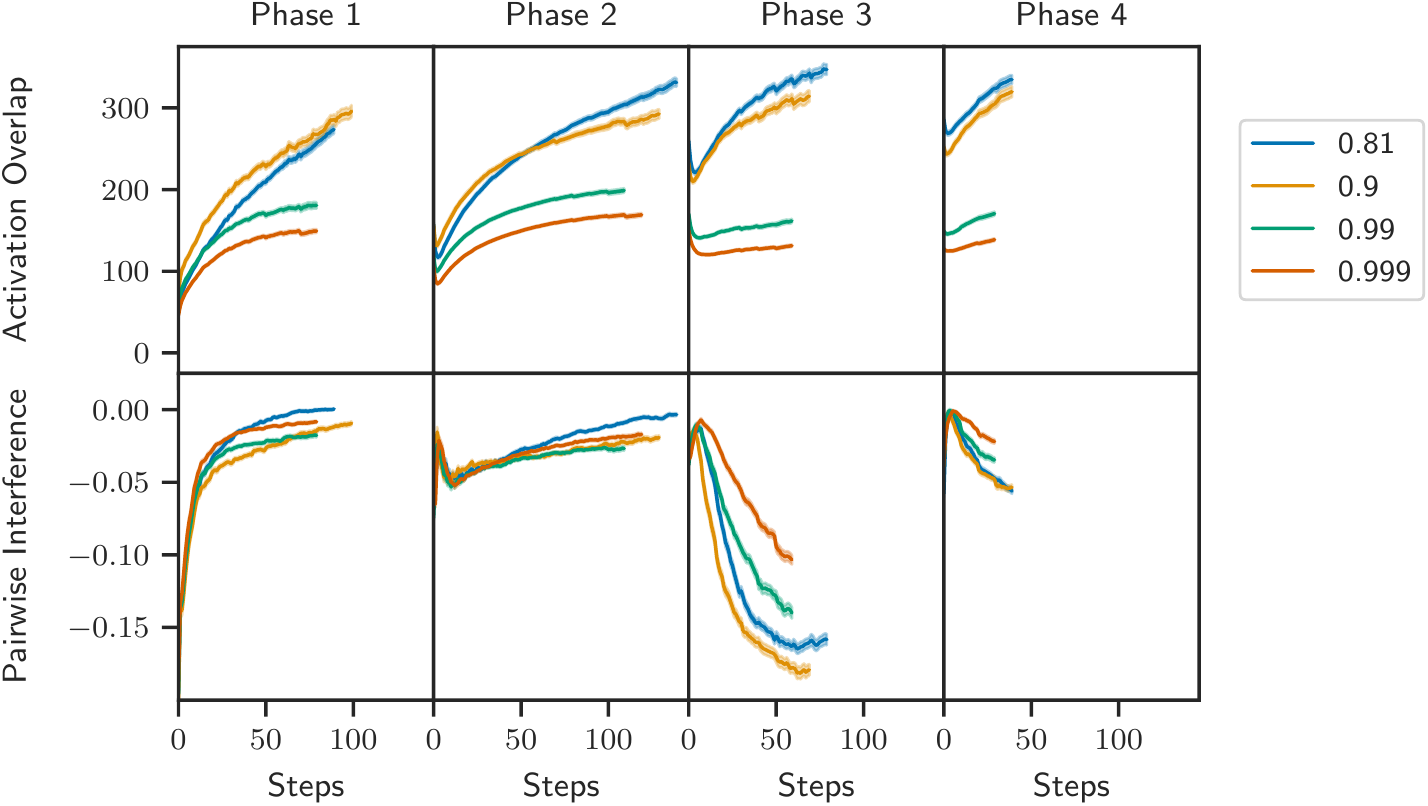}
    \caption{Activation overlap and pairwise interference in the MNIST testbed for RMSProp under different coefficients for the moving average (lower is better). Other hyperparameters were set to be consistent with Figure~\ref{fig:mnist_and_fashion_mnist_additional_interference}. Lines are averages of all runs currently in that phase and are only plotted for steps where at least half of the runs for a given optimizer are still in that phase. Standard error is shown with shading but is very small.}
    \label{fig:mnist_additional_interference_rms}
\end{figure}

\begin{figure}
    \centering
    \includegraphics[scale=0.9]{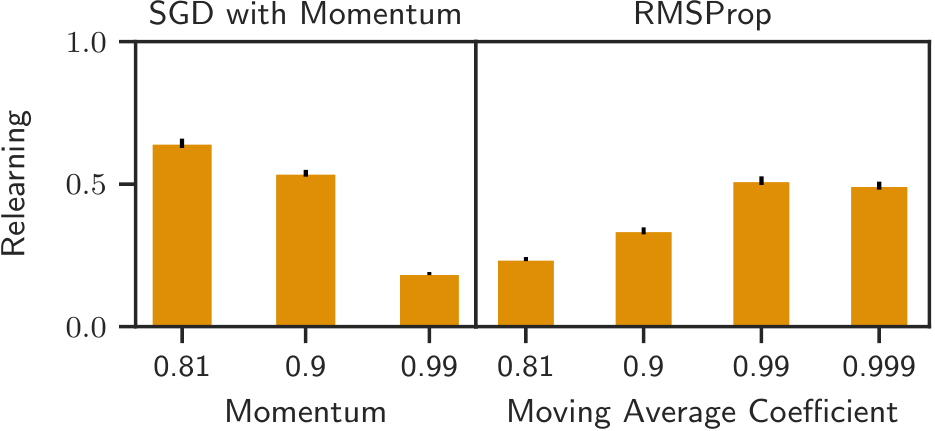}
    \caption{Relearning in the Fashion MNIST testbed for SGD with Momentum under different values of momentum, and RMSProp under different coefficients for the moving average (higher is better). Other hyperparameters were set to be consistent with Figure~\ref{fig:mnist_and_fashion_mnist_interference}. Retention in all cases was almost nonexistent and so is not reported here.}
    \label{fig:fashion_mnist_interference_momentum_and_rms}
\end{figure}

\begin{figure}
    \centering
    \includegraphics[scale=0.9]{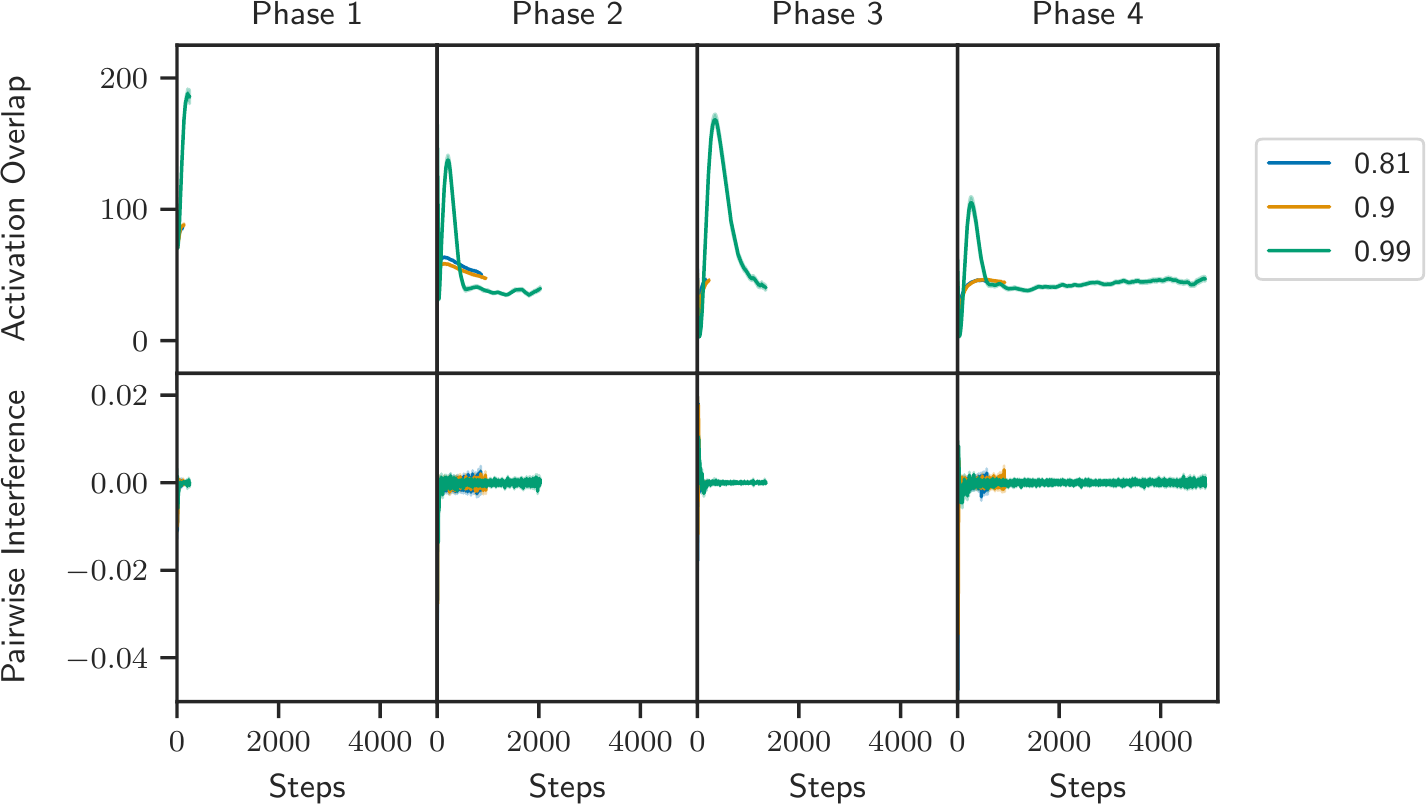}
    \caption{Activation overlap and pairwise interference in the Fashion MNIST testbed for SGD with Momentum under different values of momentum (lower is better). Other hyperparameters were set to be consistent with Figure~\ref{fig:mnist_and_fashion_mnist_additional_interference}. Lines are averages of all runs currently in that phase and are only plotted for steps where at least half of the runs for a given optimizer are still in that phase. Standard error is shown with shading but is very small.}
    \label{fig:fashion_mnist_additional_interference_momentum}
\end{figure}

\begin{figure}
    \centering
    \includegraphics[scale=0.9]{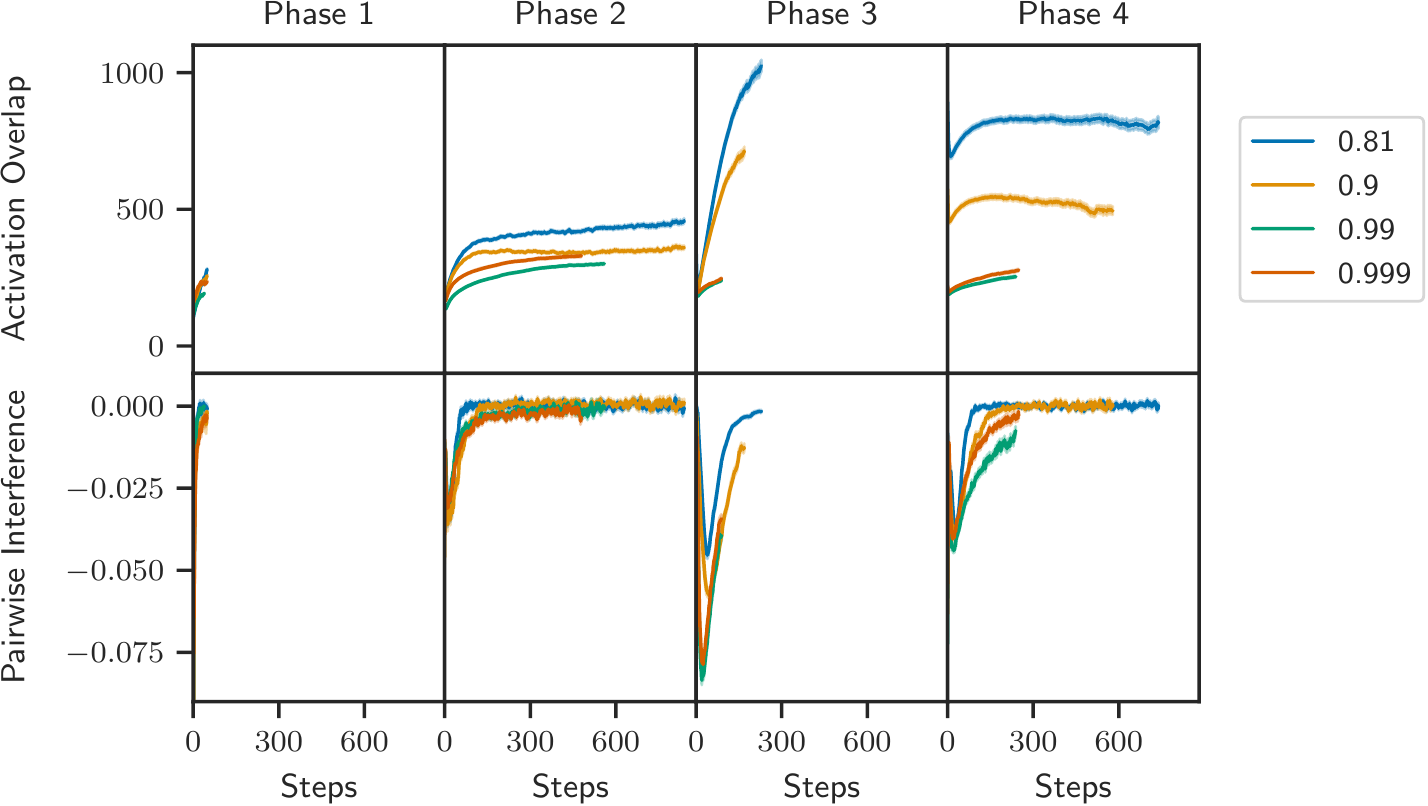}
    \caption{Activation overlap and pairwise interference in the Fashion MNIST testbed for RMSProp under different coefficients for the moving average (lower is better). Other hyperparameters were set to be consistent with Figure~\ref{fig:mnist_and_fashion_mnist_additional_interference}. Lines are averages of all runs currently in that phase and are only plotted for steps where at least half of the runs for a given optimizer are still in that phase. Standard error is shown with shading but is very small.}
    \label{fig:fashion_mnist_additional_interference_rms}
\end{figure}

\begin{figure}
    \centering
    \includegraphics[scale=0.9]{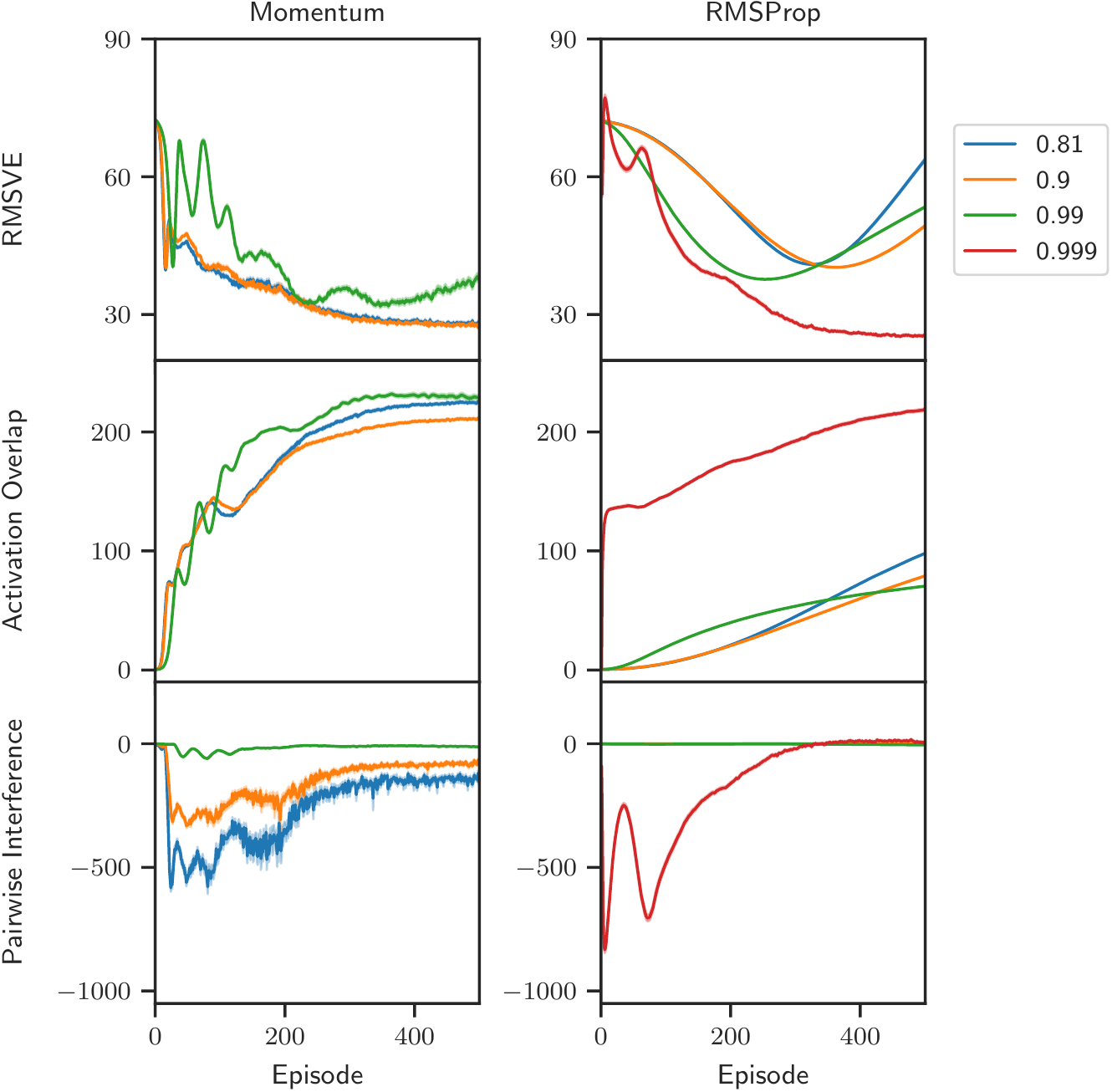}
    \caption{Performance, activation overlap, and pairwise interference in the Mountain Car testbed for SGD with Momentum under different values of momentum, and RMSProp under different coefficients for the moving average (lower is better). Other hyperparameters were set to be consistent with Figure~\ref{fig:mountain_car_and_acrobot_interference}. Standard error is shown with shading but is very small.}
    \label{fig:mountain_car_momentum_and_rms}
\end{figure}

\begin{figure}
    \centering
    \includegraphics[scale=0.9]{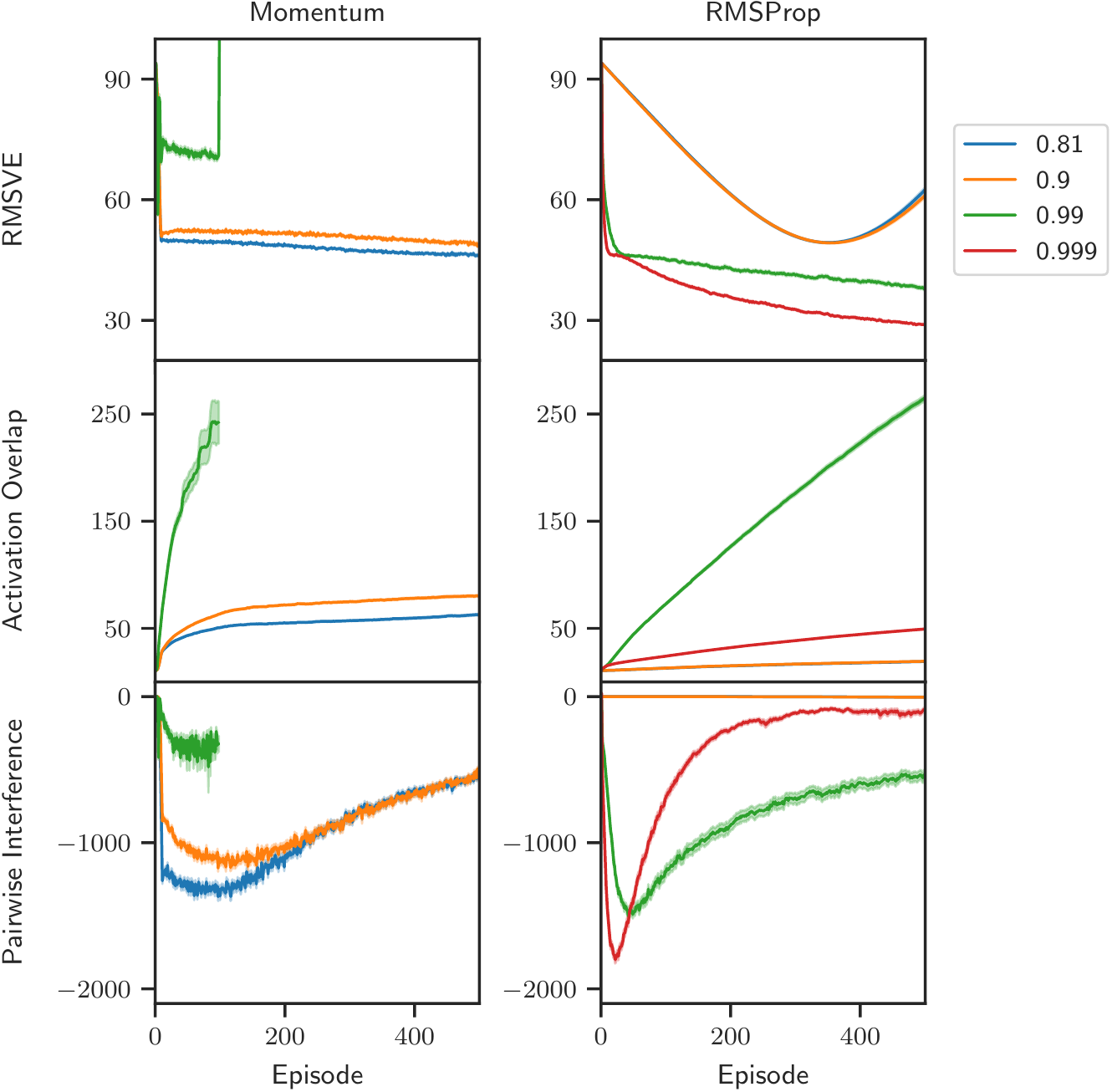}
    \caption{Performance, activation overlap, and pairwise interference in the Acrobot testbed for SGD with Momentum under different values of momentum, and RMSProp under different coefficients for the moving average (lower is better). Other hyperparameters were set to be consistent with Figure~\ref{fig:mountain_car_and_acrobot_interference}. Standard error is shown with shading but is very small.}
    \label{fig:acrobot_momentum_and_rms}
\end{figure}

\begin{figure}
    \centering
    \includegraphics[scale=0.9]{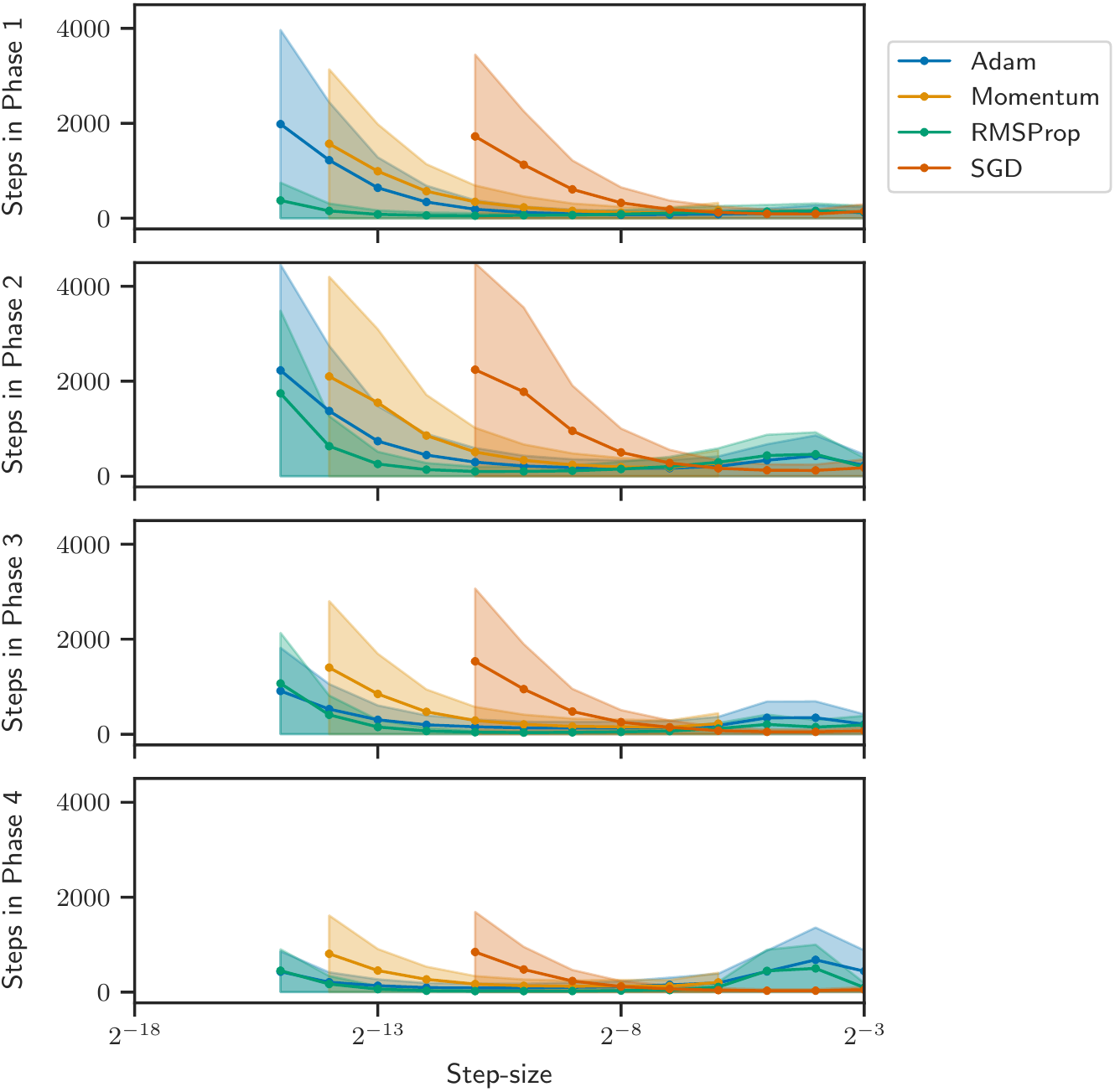}
    \caption{Number of steps needed to complete each phase in the MNIST testbed for each of the four optimizers as a function of $\alpha$ (lower is better). Other hyperparameters were set as they were in Table~\ref{tab:mnist_speed}. Lines are averages of all runs, and standard error is shown with shading. Lines are only drawn for values of $\alpha$ in which no run under the optimizer resulted in numerical instability.}
    \label{fig:mnist_speed_step-size}
\end{figure}

\begin{figure}
    \centering
    \includegraphics[scale=0.9]{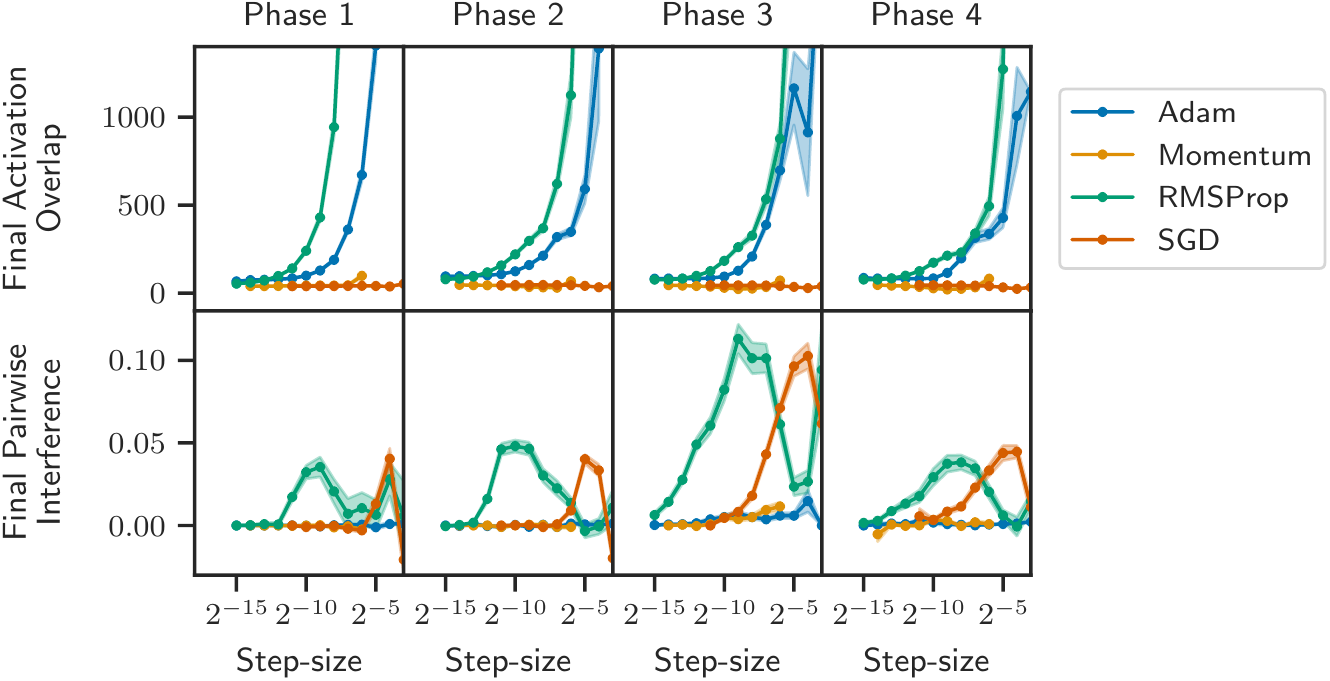}
    \caption{Final activation overlap and pairwise interference in the MNIST testbed for each of the four optimizers as a function of $\alpha$ (lower is better). Other hyperparameters were set as they were in Figure~\ref{fig:mnist_and_fashion_mnist_additional_interference}. Lines are averages of all runs, and standard error is shown with shading. Lines are only drawn for values of $\alpha$ in which no run under the optimizer resulted in numerical instability.}
    \label{fig:mnist_additional_interference_step-size}
\end{figure}

\begin{figure}
    \centering
    \includegraphics[scale=0.9]{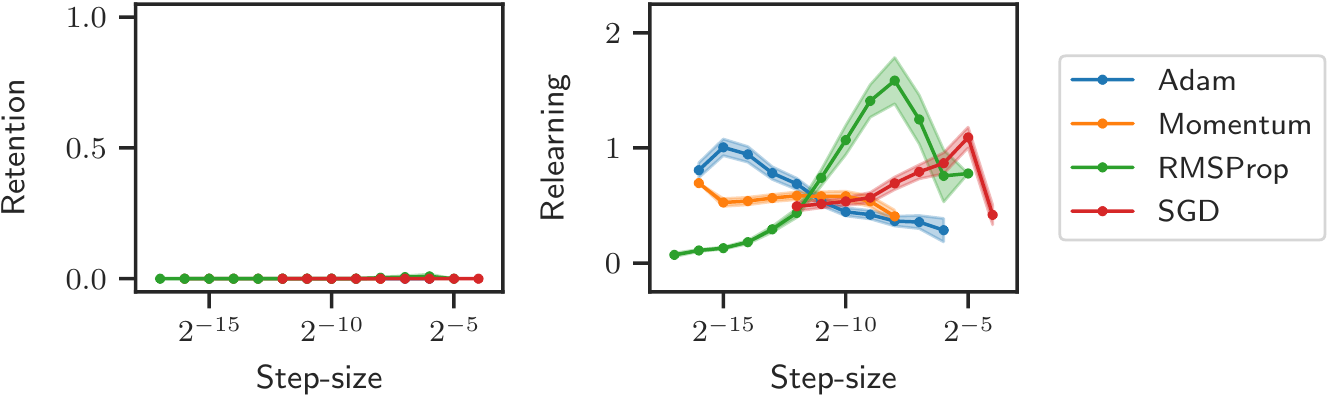}
    \caption{Retention and relearning in the Fashion MNIST testbed for each optimizer under different values of $\alpha$ (higher is better). Other hyperparameters were set to be consistent with Figure~\ref{fig:mnist_and_fashion_mnist_interference}. Lines are averages of all runs, and standard error is shown with shading. Lines are only drawn for values of $\alpha$ in which no run under the optimizer resulted in numerical instability.}
    \label{fig:fashion_mnist_interference_step-size}
\end{figure}

\begin{figure}
    \centering
    \includegraphics[scale=0.9]{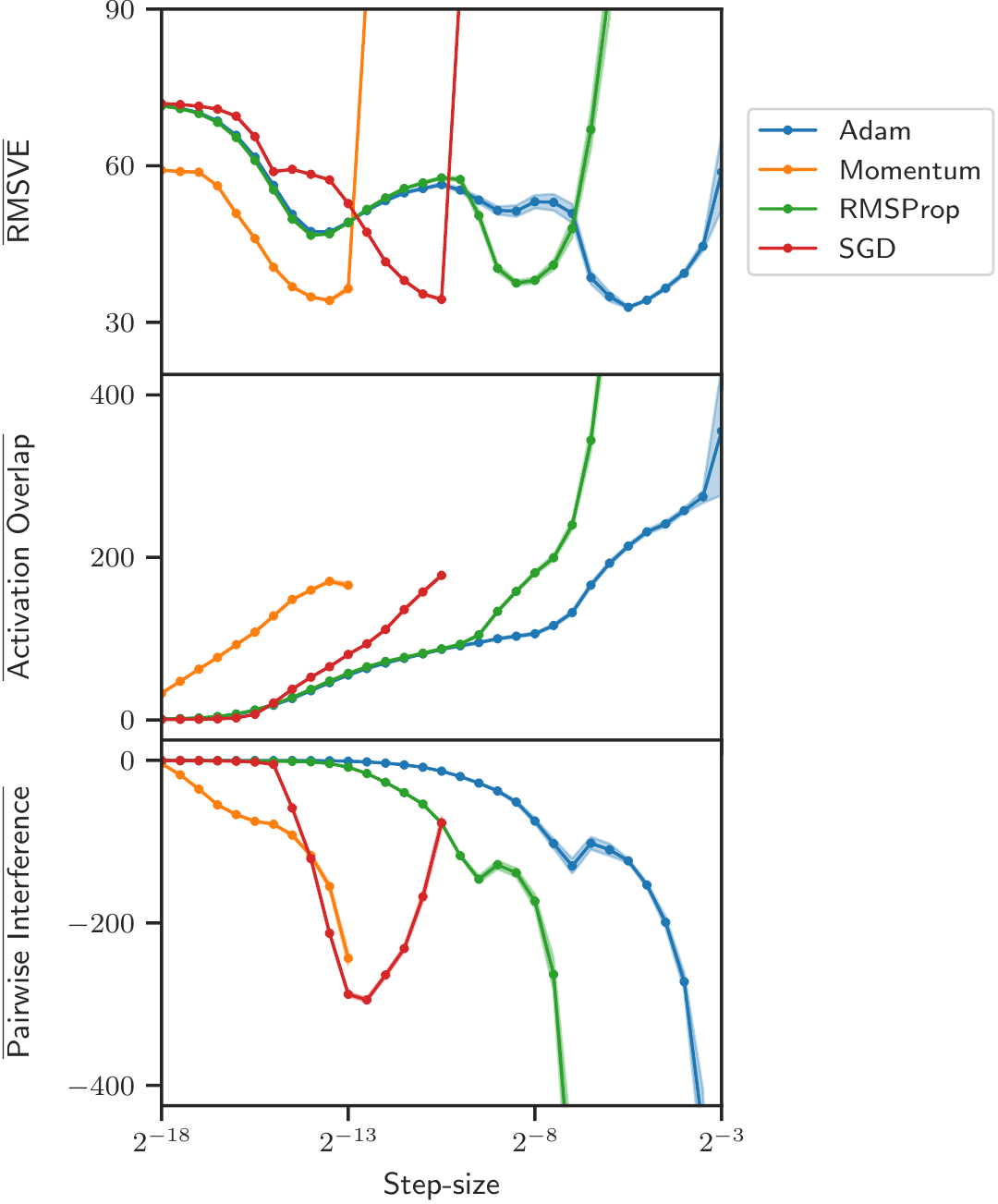}
    \caption{Mean performance and interference metrics in the Mountain Car testbed for each of the four optimizers as a function of $\alpha$ (lower is better). Other hyperparameters were set as they were in Figure~\ref{fig:mountain_car_and_acrobot_interference}. Lines are averages of all runs, and standard error is shown with shading. Both SGD and SGD with Momentum encountered numerical instability issues with certain values of $\alpha$. Lines for activation overlap and pairwise interference are drawn so as to exclude these values.}
    \label{fig:mountain_car_step-size}
\end{figure}

\begin{figure}
    \centering
    \includegraphics[scale=0.9]{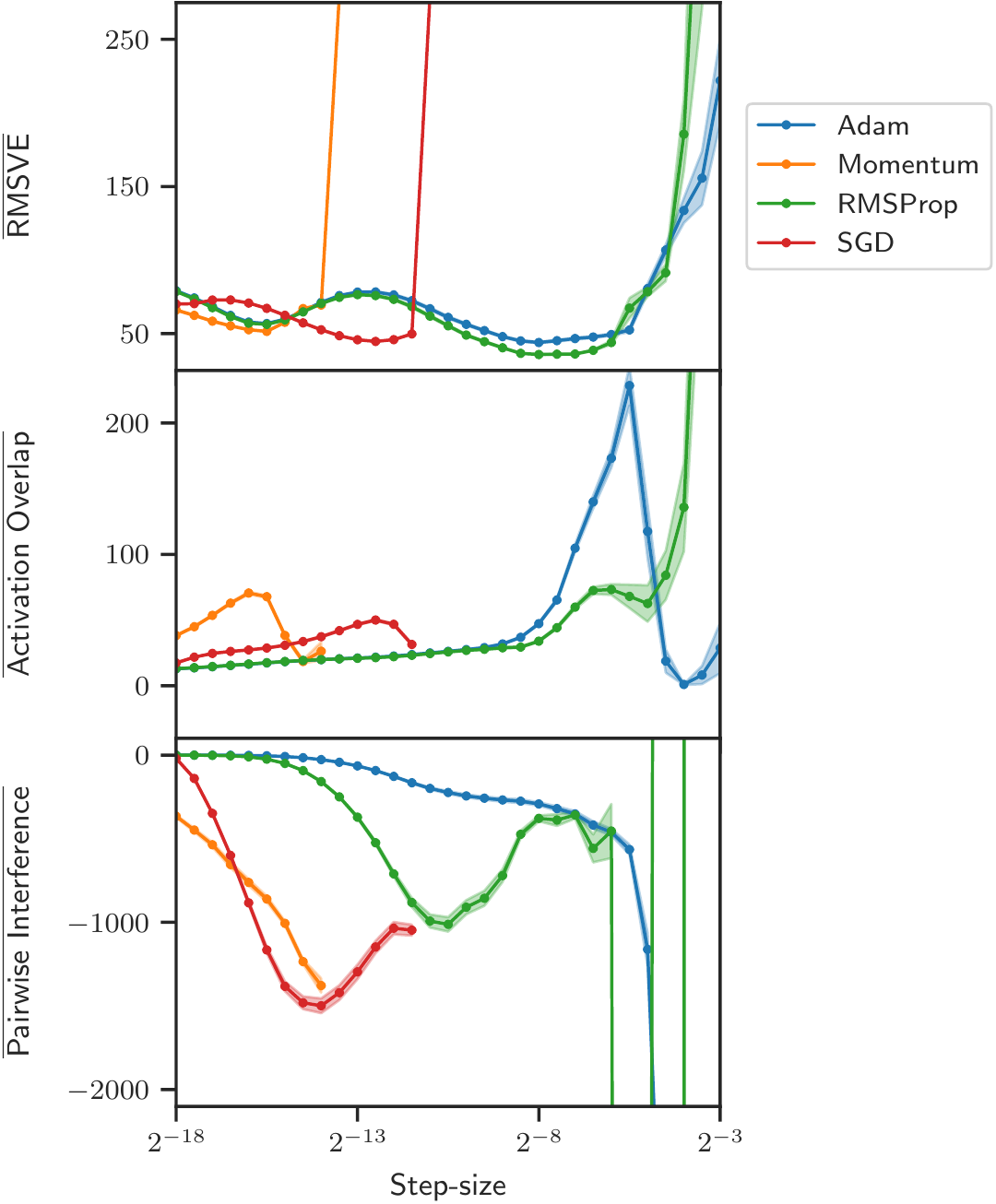}
    \caption{Mean performance, activation overlap, and pairwise interference in the Acrobot testbed for each of the four optimizers as a function of $\alpha$ (lower is better). Other hyperparameters were set as they were in Figure~\ref{fig:mountain_car_and_acrobot_interference}. Lines are averages of all runs, and standard error is shown with shading. Both SGD and SGD with Momentum encountered numerical instability issues with certain values of $\alpha$. Lines for activation overlap and pairwise interference are drawn so as to exclude these values.}
    \label{fig:acrobot_step-size}
\end{figure}

Figures \ref{fig:mnist_additional_interference_momentum} and \ref{fig:mnist_additional_interference_rms} show both the activation overlap and pairwise interference in the MNIST testbed under different coefficients for the moving average in RMSProp, and values of the momentum parameter in SGD with Momentum, respectively. Figures \ref{fig:fashion_mnist_additional_interference_momentum} and \ref{fig:fashion_mnist_additional_interference_rms} show the same in the Fashion MNIST testbed. Tables \ref{tab:mnist_speed_momentum_and_rms} and \ref{tab:fashion_mnist_speed_momentum_and_rms} then show the corresponding variations in performance for the two testbeds. Note than the data in Table~\ref{tab:mnist_speed_momentum_and_rms} was generated with the other hyperparameters being the same as in Table~\ref{tab:mnist_speed}, and the data in Table~\ref{tab:fashion_mnist_speed_momentum_and_rms} was generated with the other hyperparameters being the same as in Table~\ref{tab:fashion_mnist_speed}. Figures \ref{fig:mountain_car_momentum_and_rms} and \ref{fig:acrobot_momentum_and_rms} show the results of the same perturbations on the performance, activation overlap, and pairwise interference in the Mountain Car and Acrobot testbeds, respectively. Similarly, Figures \ref{fig:mnist_speed_step-size} and \ref{fig:mnist_additional_interference_step-size}, \ref{fig:mountain_car_step-size}, and \ref{fig:acrobot_step-size} show the performance as well as the activation overlap, and pairwise interference as a function of $\alpha$ in the MNIST, Mountain Car, and Acrobot testbeds, respectively. Figure~\ref{fig:fashion_mnist_interference_momentum_and_rms} shows the variations in retention and relearning in the Fashion MNIST setting for different coefficients for the moving average in RMSProp, as well as values of the momentum parameter in SGD with Momentum. Note that retention was almost nonexistent in all cases. Finally, Figure~\ref{fig:fashion_mnist_interference_step-size} shows the same but as a function of $\alpha$.

As with the results in the main text, there is a clear relationship between the momentum parameter in SGD with Momentum and catastrophic forgetting, as well as between the coefficients for the moving average in RMSProp and catastrophic forgetting. However, in most instances, the relationship is smooth, with small variations in the parameter producing small variations in the amount of catastrophic forgetting observed. Some discrepancies can be observed here (e.g., activation overlap under SGD with Momentum in Figure~\ref{fig:mountain_car_momentum_and_rms}), though they remain in the minority.

Regarding the effect of $\alpha$ on catastrophic forgetting, the results here are again consistent with the conclusions reached in the main text. Namely, that--like with the momentum parameter in SGD with Momentum and the coefficients for the moving average in RMSProp--there is a pronounced relationship between the amount of catastrophic forgetting observed and the value of $\alpha$. Again, though, this relationship is smooth with similar values of $\alpha$ producing similar amounts of forgetting.

\begin{table}
    \caption{Average number of steps to complete each phase in the MNIST testbed for SGD with Momentum under different values of momentum, and RMSProp under different coefficients for the moving average.}
    \label{tab:mnist_speed_momentum_and_rms}
    \centering
    \begin{tabular}{lccccc}
        \toprule
        Optimizer & Gamma & Phase 1 & Phase 2 & Phase 3 & Phase 4 \\
        \midrule
        \multirow{3}{*}{Momentum} & 0.81 & {\bf 122.49$\pm$2.66} & {\bf 155.84$\pm$2.29} & {\bf 106.80$\pm$2.23} & {\bf 72.99$\pm$1.47} \\
        & 0.9 & 135.88$\pm$2.86 & 192.18$\pm$2.38 & 155.03$\pm$2.67 & 116.55$\pm$1.90 \\
        & 0.99 & 249.62$\pm$5.10 & 542.00$\pm$6.12 & 806.30$\pm$11.76 & 885.74$\pm$14.62 \\
        \midrule
        \multirow{4}{*}{RMSProp} & 0.81 & 70.15$\pm$1.51 & 119.08$\pm$1.64 & 62.59$\pm$1.35 & 29.13$\pm$0.78 \\
        & 0.9 & 70.49$\pm$1.58 & 108.44$\pm$1.62 & 47.98$\pm$1.18 & 26.39$\pm$0.73 \\
        & 0.99 & {\bf 61.37$\pm$1.30} & {\bf 96.92$\pm$1.31} & {\bf 43.89$\pm$1.03} & {\bf 23.21$\pm$0.70} \\
        & 0.999 & {\bf 60.19$\pm$1.25} & 100.08$\pm$1.28 & 49.29$\pm$1.11 & {\bf 24.54$\pm$0.81} \\
        \bottomrule
    \end{tabular}
\end{table}

\begin{table}
    \caption{Average number of steps to complete each phase in the Fashion MNIST testbed for SGD with Momentum under different values of momentum, and RMSProp under different coefficients for the moving average.}
    \label{tab:fashion_mnist_speed_momentum_and_rms}
    \centering
    \begin{tabular}{lccccc}
        \toprule
        Optimizer & Gamma & Phase 1 & Phase 2 & Phase 3 & Phase 4 \\
        \midrule
        \multirow{3}{*}{Momentum} & 0.81 & {\bf 85.12$\pm$1.88} & {\bf 696.52$\pm$10.37} & {\bf 139.80$\pm$1.42} & {\bf 471.78$\pm$9.75} \\
        & 0.9 & 107.46$\pm$2.17 & 815.92$\pm$9.61 & 205.26$\pm$1.74 & 754.86$\pm$10.88 \\
        & 0.99 & 189.92$\pm$4.00 & 1780.96$\pm$20.49 & 1127.34$\pm$17.52 & 4116.42$\pm$59.58 \\
        \midrule
        \multirow{4}{*}{RMSProp} & 0.81 & 39.13$\pm$0.99 & 644.30$\pm$12.44 & 187.68$\pm$2.89 & 569.92$\pm$12.34 \\
        & 0.9 & {\bf 38.17$\pm$1.04} & 620.00$\pm$12.86 & 139.40$\pm$2.90 & 393.58$\pm$12.08 \\
        & 0.99 & {\bf 36.63$\pm$0.91} & 421.16$\pm$8.70 & {\bf 79.36$\pm$1.16} & {\bf 186.82$\pm$7.07} \\
        & 0.999 & {\bf 36.39$\pm$0.94} & {\bf 375.22$\pm$7.84} & {\bf 80.90$\pm$1.18} & {\bf 186.18$\pm$6.33} \\
        \bottomrule
    \end{tabular}
\end{table}

\begin{figure}
    \centering
    \includegraphics[scale=0.9]{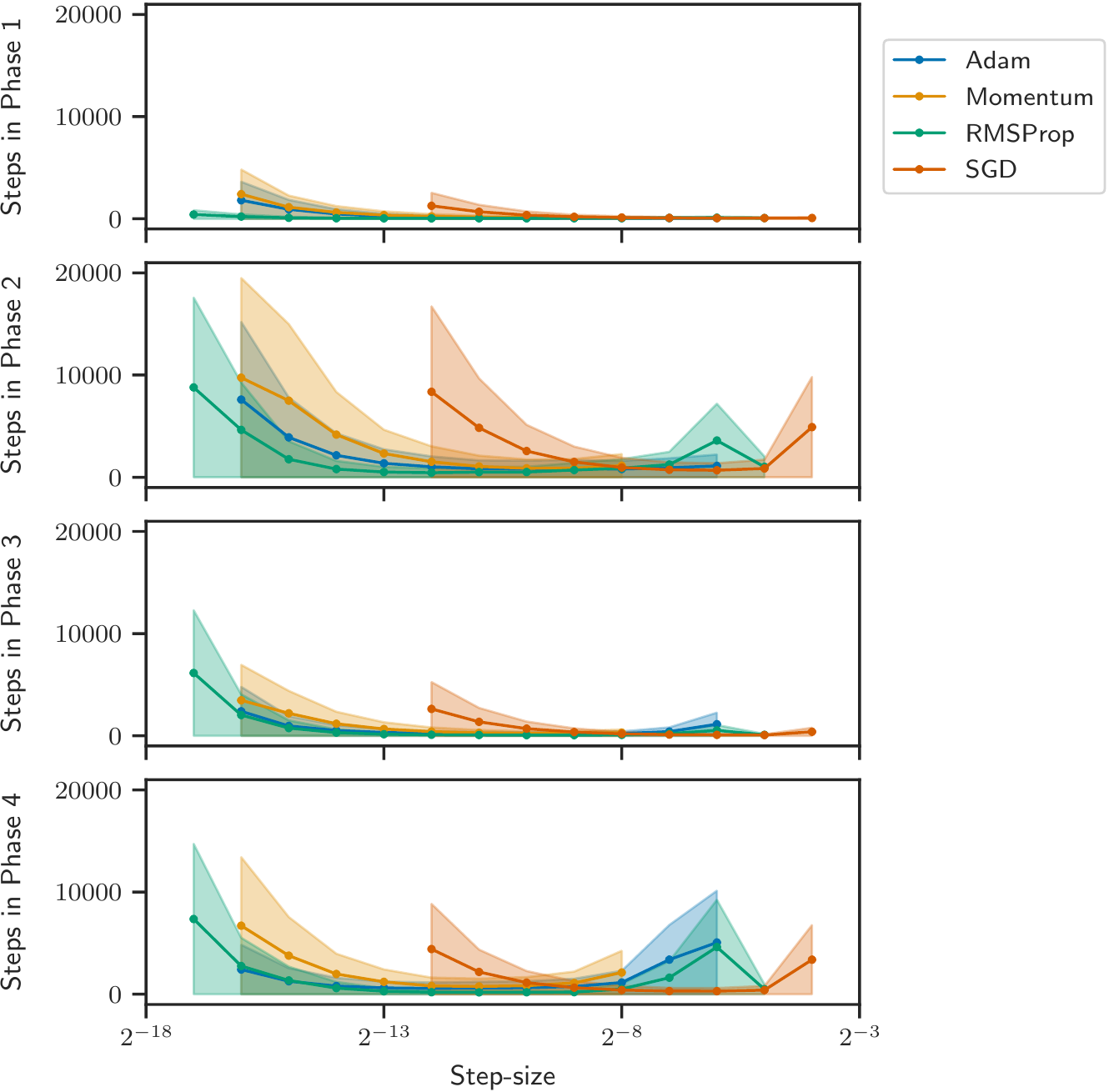}
    \caption{Number of steps needed to complete each phase in the Fashion MNIST testbed for each of the four optimizers as a function of $\alpha$ (lower is better). Other hyperparameters were set as they were in Table~\ref{tab:fashion_mnist_speed}. Lines are averages of all runs, and standard error is shown with shading. Lines are only drawn for values of $\alpha$ in which no run under the optimizer resulted in numerical instability.}
    \label{fig:fashion_mnist_speed_step-size}
\end{figure}

\begin{figure}
    \centering
    \includegraphics[scale=0.9]{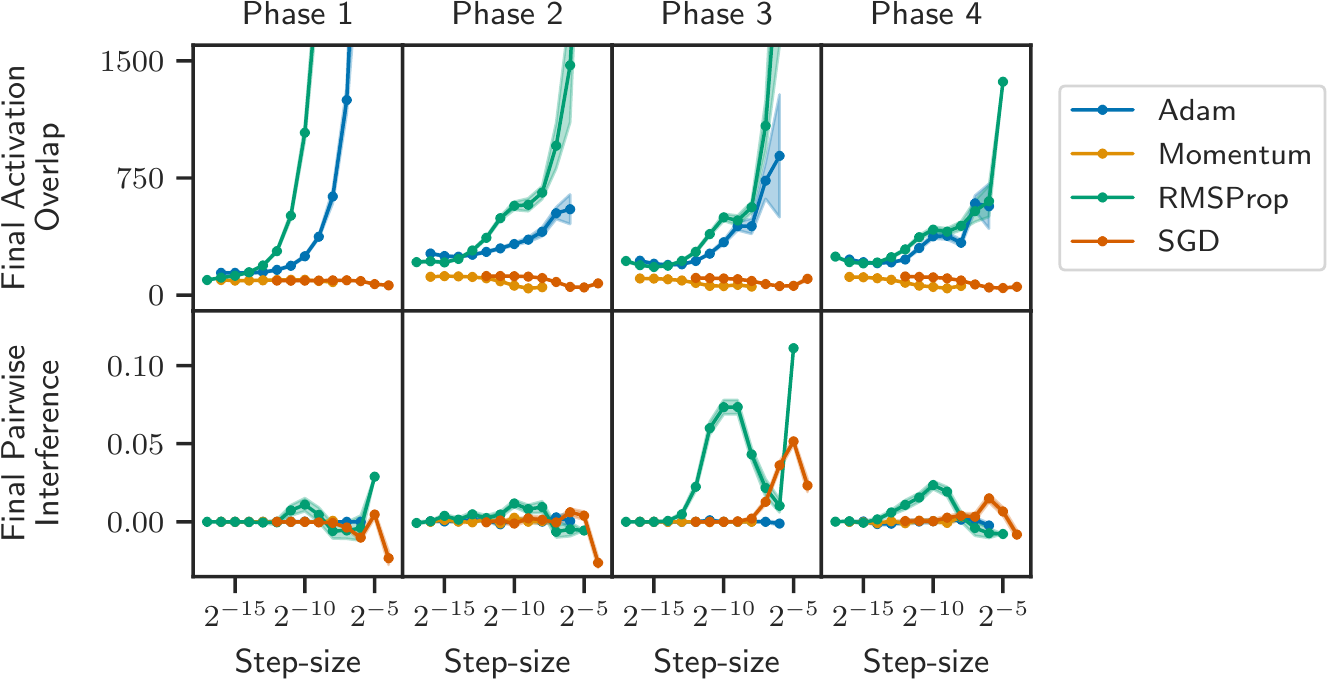}
    \caption{Final activation overlap and pairwise interference in the Fashion MNIST testbed for each of the four optimizers as a function of $\alpha$ (lower is better). Other hyperparameters were set as they were in Figure~\ref{fig:mnist_and_fashion_mnist_additional_interference}. Lines are averages of all runs, and standard error is shown with shading. Lines are only drawn for values of $\alpha$ in which no run under the optimizer resulted in numerical instability.}
    \label{fig:fashion_mnist_additional_interference_step-size}
\end{figure}

\end{document}